\documentclass{article}

\PassOptionsToPackage{numbers, compress}{natbib}
\usepackage[preprint]{neurips_2021}

\usepackage[utf8]{inputenc} 
\usepackage[T1]{fontenc}    
\usepackage{hyperref}       
\usepackage{url}            
\usepackage{booktabs}       
\usepackage{amsfonts}       
\usepackage{nicefrac}       
\usepackage{microtype}      
\usepackage{xcolor}         

\usepackage{amsmath}
\usepackage{tikz}
\usepackage{multirow, bigdelim} 
\usepackage{graphicx} 
\usepackage{pgfplots}
\usepackage{amssymb}
\usepackage{tabularx, blkarray}
\usepackage{array}
\usepackage[ruled, vlined]{algorithm2e}
\usepackage{mathtools}

\pgfplotsset{compat=1.14}
\usetikzlibrary{shapes.geometric}
\usetikzlibrary{arrows.meta}
\DeclareMathOperator{\EX}{\mathbb{E}}
\DeclareMathOperator{\loss}{\mathcal{L}} 
\newcommand{\rotate}[1]{\rotatebox{90}{#1}} 

\title{Points2Polygons: Context-Based Segmentation from Weak Labels Using Adversarial Networks}

\author{%
  Kuai Yu  \\
  Esri\\
  \texttt{dyu@esri.com} \\
  \And
  Hakeem Frank \\
  Esri \\
  \texttt{hfrank@esri.com} \\
  \AND
  Daniel Wilson \\
  SeerAI \\
  \texttt{dwilson@seerai.space} \\
}

\begin{document}

\bibliographystyle{abbrvnat}

\maketitle

\begin{abstract}
  In applied image segmentation tasks, the ability to provide numerous and precise 
  labels for training is paramount to the accuracy of the model at inference time. 
  However, this overhead is often neglected, and recently proposed segmentation 
  architectures rely heavily on the availability and fidelity of ground truth 
  labels to achieve state-of-the-art accuracies. Failure to acknowledge the 
  difficulty in creating adequate ground truths can lead to an over-reliance 
  on pre-trained models or a lack of adoption in real-world applications. We 
  introduce Points2Polygons (P2P), a model which makes use of contextual metric 
  learning techniques that directly addresses this problem. Points2Polygons 
  performs well against existing fully-supervised segmentation baselines with 
  limited training data, despite using lightweight segmentation models (U-Net 
  with a ResNet18 backbone) and having access to only weak labels in the form 
  of object centroids and no pre-training. We demonstrate this on several different 
  small but non-trivial datasets. We show that metric learning using contextual 
  data provides key insights for self-supervised tasks in general, and allow 
  segmentation models to easily generalize across traditionally label-intensive 
  domains in computer vision.
\end{abstract}

\section{Introduction}

Segmentation tasks based on remotely-sensed imagery rely heavily on the availability of good quality semantic or 
polygon labels to be useful \citep{wang:weaklysupervised2020, zhu:pickqualityeval2019}. 
Specifically, \cite{ahn:semanticaffinity2018} notes that the performance of ConvNets for segmentation tasks
is largely dependent on the amount of time spent creating training labels. These tasks can be as simple as instance counting 
\cite{li:oilpalm2017,neupane:bananacount2019, osco:citruscount2020,yao:treecount2021} or object identification \citep{cha:crackdetection2017,gulgec:damagedetect2017}. 
However, in practice these labels tend not to be readily available outside of established tasks 
such as building footprint extraction or vehicle detection, so it is often necessary 
to spend significant manual effort to digitize these polygons \citep{wang:weaklysupervised2020}. 
This drastically increases the cost to build segmentation models for practical applications, and 
presents a major barrier for smaller organizations. 

Several approaches have been proposed in response to this issue such as weak-label learning 
\citep{ahn:interpixel2019,laradji:pointsuper2019,paul:domainadapt2020, wang:weaklysupervised2020} 
and self-supervision via a contrastive objective in Siamese or Triplet networks 
\citep{caron:emerging2021, dhere:selfsupseg2021, jaiswal:surveyselfsup2020,leyva:contrastopt2021}. 
However, these approaches come with their own problems. Models based on weak 
labels for instance are challenging to evaluate since ground truth polygons 
are not necessarily meaningful labels.
In the case of aerial imagery, a model which fixates on salient features of an object such as a 
chimney or a wind shield should not be penalized for having a non-intuitive interpretation of what 
best defines a building or a vehicle. 
 
This paper contributes a novel approach to semantic segmentation with weak labels by incorporating 
constraints in the segmentation model $S$ through an adversarial objective, 
consisting of an object localization discriminator $D_1$ and a contextual discriminator $D_2$. 

In our framework, $D_1$ discriminates between the true image $I_R$ and a fake 
image $I_{F_1}$, where $I_{F_1}$ is a spatially similar sample in the training set with 
the positive object superimposed to it. By treating the segmentation model as 
a generator, and implementing an additional discriminator, we propagate learnable 
gradients back to the generator. This allows the segmentation model to produce masks 
which when used to superimpose the original image into neighboring contexts, produce 
believable “generated” images. The full formulation of the discriminator and the superimpose function 
$f$ is discussed in Section \ref{subsection:adversarial}. $D_2$ is an additional contextual discriminator 
which we introduce in order to constrain artifacts that arise from having only one discriminator.

We show that this method produces high quality segmentations without having to consider imagery 
gradients and uses only buffered object polygon centroids as inputs. The major contribution of this paper is a new form of
semi-supervised segmentation model that can be applied to segment objects using weak labels, such as those 
provided by manual annotation or object detection models which only provide bounding boxes. Future extensions 
of this work will address the localization task, providing end-to-end instance segmentation from weak labels and 
low requirements for the number of annotations to reach acceptable performance. 

\section{Related Work}

\begin{figure}[!t]
  \begin{center}
  \setlength{\tabcolsep}{1pt}
    \begin{tabular}{ccccccc}
  \includegraphics[width=.14\linewidth]{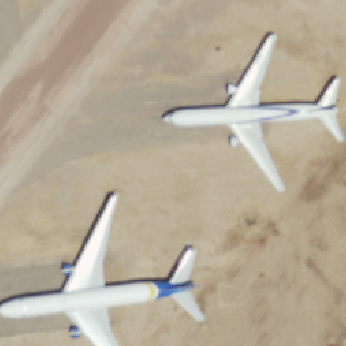} &
   \includegraphics[width=.14\linewidth]{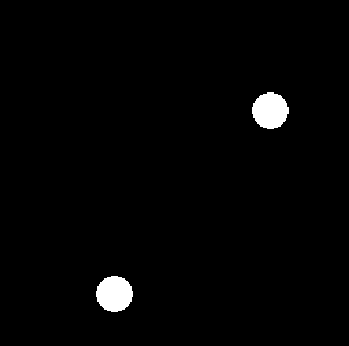} &
   \includegraphics[width=.14\linewidth]{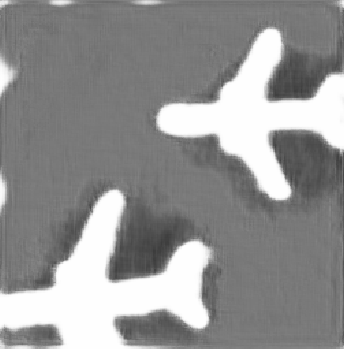} &
   \includegraphics[width=.14\linewidth]{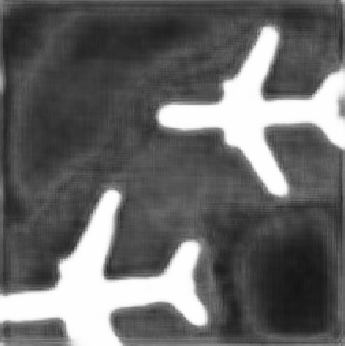} &
   \includegraphics[width=.14\linewidth]{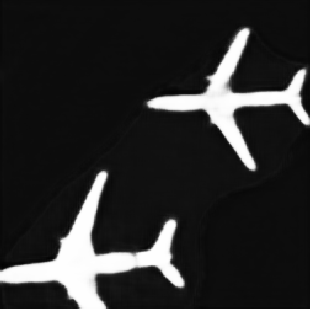} & 
   \includegraphics[width=.14\linewidth]{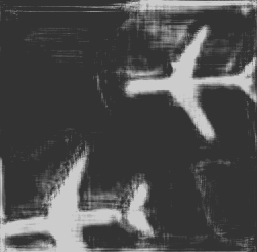} &
   \includegraphics[width=.14\linewidth]{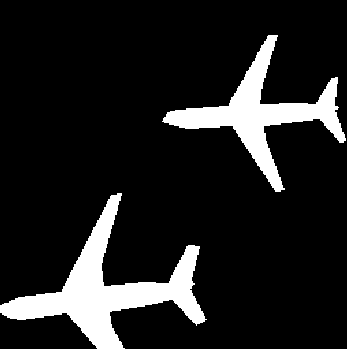} \\
   \includegraphics[width=.14\linewidth]{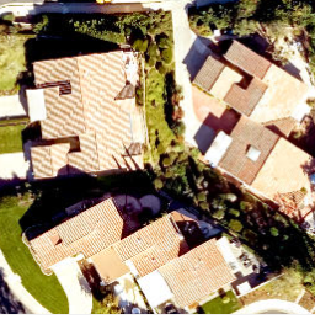} &
   \includegraphics[width=.14\linewidth]{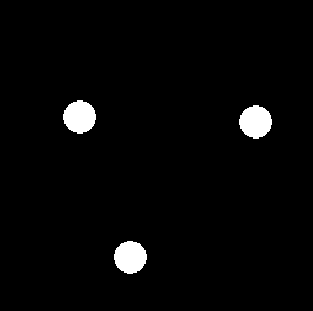} &
   \includegraphics[width=.14\linewidth]{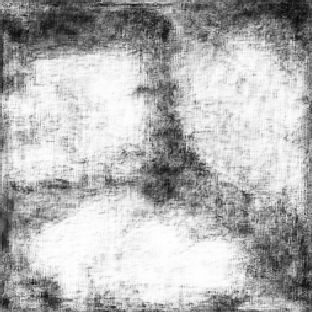} & 
   \includegraphics[width=.14\linewidth]{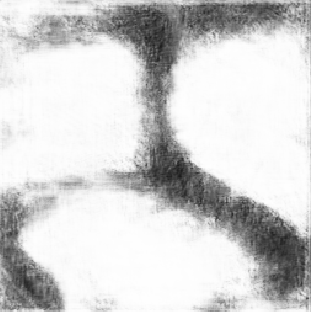} & 
   \includegraphics[width=.14\linewidth]{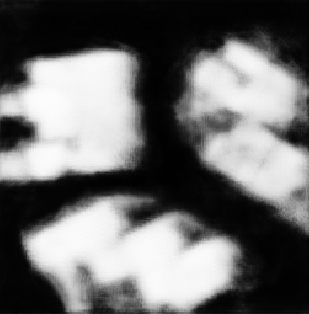} & 
   \includegraphics[width=.14\linewidth]{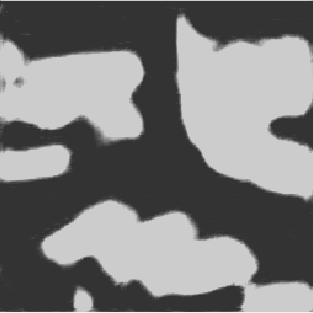} &
   \includegraphics[width=.14\linewidth]{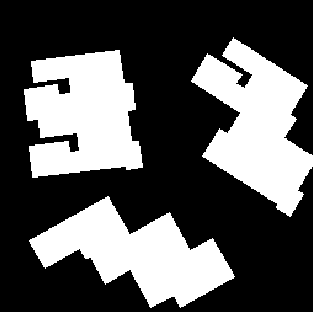} \\
   (a) & (b)  & (c) & (d) & (e) & (f)  & (g)  \\
  \end{tabular}
  
  \end{center}
  \caption[Predicted segmentation masks from the Aircraft and Woolsey datasets]{
    Predicted segmentation masks from the Aircraft and Woolsey datasets using fully-supervised
    UNet models with ResNet backbones (c-e) and our semi-supervised P2P model (f). All generated masks
    come from training with Dice loss, besides (e), which is trained with cross entropy loss for comparison. 
    (a) Input image to each model. (b) Pseudo label $\tilde{y}$ used to train P2P. (c) Output from 
    a UNet-101 baseline model. (d) Output from UNet-50. (e) Output from UNet-50 trained with cross-entropy. 
    (f) Output from P2P (ours). (g) Ground Truth labels.
  \label{fig:bangers} }
  
\end{figure}

Points2Polygons shares ideas in common with four general types of segmentation models which 
we will discuss here:

\paragraph{Footprint extraction models}

We make a reference to building footprint extraction models because these models 
aim to extract (either via semantic or instance segmentation) complex geometries 
that come from a much more complex distribution space - in terms of shape and texture variance -  
than features such 
as vehicles or trees \citep{bittner:footprintdsm2016, chawda:extractfootprints2018,zhu:mapnet2019}. 
Building footprints are also a class of features which are 
conditionally dependent on contextual information; a building is much more likely 
to be found in a suburban context next to trees or driveways as opposed to in 
the desert. 

Existing approaches however rely heavily on segmentation labels
\citep{chawda:extractfootprints2018, zhu:mapnet2019} and primarily focus on improving 
the accuracy from a supervised learning approach given large amounts of 
training data, such as in the SpaceNet \citep{vanetten:spacenet2018} and WHU \citep{xia:dota2017} datasets. 
This is unfortunately not readily available for many worthwhile segmentation tasks 
such as well pad extraction for localizing new fracking operations or airplane detection 
for civil and remote sensing applications. 

\paragraph{Semi-supervised or unsupervised methods}

Other methods such as \citep{ahn:semanticaffinity2018, ahn:interpixel2019, laradji:pointsuper2019, paul:domainadapt2020,wang:weaklysupervised2020}
and \citep{dai:boxsup2015} attempt to address the scarcity of existing ground truth labels by 
adopting a semi-supervised or unsupervised approach. These include methods which 
attempt to perform segmentation based on a single datapoint as demonstrated in \cite{wang:weaklysupervised2020}. 
However approaches such as these only factor in low-frequency information and are designed for 
landcover classification. \citet{laradji:pointsuper2019} utilizes point-based 
pseudo labels to segment objects of interest in imagery using a network with a localization
branch and an embedding branch to generate full segmentation masks. While similar in nature to P2P,
\citep{laradji:pointsuper2019} optimizes a squared exponential distance metric between same-class 
pixels in embedding space to generate segmentation masks, whereas P2P frames the segmentation task
as a simple image-to-image translation problem with localization constraints 
enforced by an adversarial objective. Our approach thus circumvents the need for pair-wise 
calculations from a similarity function with the additional benefit 
of a separately tunable constraint. 

\cite{NEURIPS2019_32bbf7b2} attempts to solve segmentation problems in a completely unsupervised setting using
an adversarial architecture to segment regions that can be clipped and redrawn using a generator network such that
the generated images is aligned with the original. P2P goes further by using easy-to-obtain supervised labels that 
consist of points within individual instances. 

An alternate approach to weak supervision for segmentation is to utilize image-level 
class labels in conjunction with class attention maps \citep{ahn:interpixel2019,oquab:freelocalization2015}.
Besides the low labeling cost, the benefit of this approach is that the class 
boundary for the object of interest is informed by contextual information in the rest of the image, in contrast 
to pixel-level supervision, where the model is encouraged to learn shape and 
color cues in the object of interest. 
P2P incorporates image-level information by learning characteristics of an object's context via 
a second contextual discriminator, $D_2$, as described in Section \ref{subsection:adversarial}.

\paragraph{Adversarial methods} 

One of the primary difficulties of learning segmentation from weak labels is 
that it is difficult to evaluate the quality of the segmentation model in the 
absence of ground-truth polygons. 
Traditional methods tackle this challenge by introducing priors or 
domain-specific constraints, such as in superpixel-based models \citep{csillik:slicremotesensing2017, fiedler:powerslic2020, hartley:superpixelcnn2019,ren:gslicr2018}. 
However this approach is difficult to generalize to cases where the landscape changes 
or in the presence of covariate shift introduced via a new dataset. 

In these cases, it may be necessary to adjust model hyperparameters, such as is commonly done in 
experiments with production-level segmentation pipelines 
in order to compensate for these new domains which in itself can be very time-intensive.

We draw inspiration from works that leverage Generative Adversarial Networks (GANs) which aim to introduce a 
self-imposed learning objective via a discriminator $D$ which models and attempts to 
delineate the difference in distribution between a real and fake input set. In particular,
\citet{luc:semseg2016} uses an adversarial network to enforce higher-order consistency
in the segmentation model such that shape and size of label regions are considered, while 
\citet{souly:semigan2017} use GANs to create additional training data. 

This is a form of self-supervision which we are able to incorporate into P2P. By introducing 
discriminators, we get past the need to impose strong human priors and instead allow 
the model to determine whether or not the segmentation output is “realistic”. 
In effect, our optimization objective enforces contextual consistency
through a $\min\max$ formulation. We additionally modify the vanilla GAN
objective in order to prevent high model bias, 
which we outline under Contextual Similarity in Section \ref{subsection:context} 

\section{Methodology}

\begin{figure}[!t]
  \centering
  \begin{tikzpicture}[scale=.70]
    \begin{scope}[main node/.style={thick,draw}]
      \node [main node](y_tilde) at (0,0) {$\tilde{y}$};
      \node [style={circle,thick,draw}](G) at (2, 0) {$S$};
      \node [main node](y_hat) at (4.25, 1) {$\hat{y}$};
      \node [main node](y_prime_c) at (4.25, -1) {$\hat{y}^c$};
      \node [style={circle,thick,draw}](D1) at (9.25, 2) {$D_1$};
      \node [style={circle,thick,draw}](D2) at (9.25, -2) {$D_2$};
      \node  [main node](I_R) at (2, 2) {$I_{R}$};

      \node [main node](I_F1) at (7.5, 1) {$I_{F_1}$};
      \node [main node](I_F2) at (7.5, -1) {$I_{F_2}$};

      \node [main node](x_bar)at (2, -2) {$I_{ctx}$};
      \node (f1) at (6, 1) {$f$};
      \node (f2) at (6, -1) {$f$};
  \end{scope}
  
  \begin{scope}[>={Stealth[black]},
                every node/.style={fill=white,circle},
                every edge/.style={draw=black,very thick}]
      \path [->] (y_tilde) edge (G);
      \path [->] (G) edge (y_hat);
      \path [->] (I_F1) edge (D1);
      \path [->] (y_hat) edge (y_prime_c);
      \path [->] (y_hat) edge (f1);
      \path [->] (f1) edge (I_F1);
      \path [->] (I_R) edge [bend left] (D1);
      \path [->] (I_R) edge (f1);
      \path [->] (I_R) edge (f2);
      \path [->] (y_prime_c) edge (f2);
      \path [->] (x_bar) edge [bend right](D2);
      \path [->] (x_bar) edge (f2);
      \path [->] (x_bar) edge (f1);
      \path [->] (f2) edge (I_F2);
      \path [->] (I_F2) edge (D2); 
      \path [->] (I_R) edge (G);
  \end{scope}
  \end{tikzpicture}
  \caption{Our modified adversarial objective. Our segmentation model 
  $S$ takes in pseudo label $\tilde{y}$ and $I_R$ and returns a predicted 
  segmentation mask $\hat{y}$, which is then used to generate fake images 
  $I_{F_1}$ and $I_{F_2}$, which are used to train the discriminators $D_1$ 
  and $D_2$. See Section \ref{subsection:adversarial} for more details on this 
  procedure.}
  \end{figure}
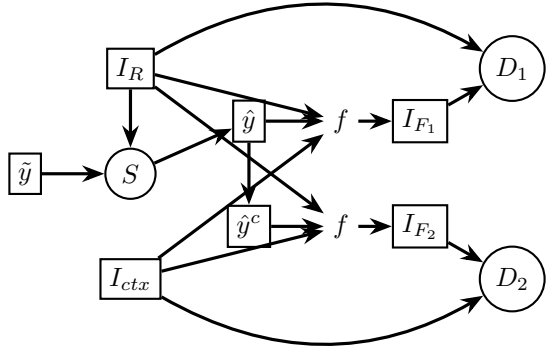

\subsection{Adversarial Objective}\label{subsection:adversarial}

This is the vanilla GAN objective:

\begin{equation}\label{gan_objective}
  \min_{G} \max_{D} \loss(G, D) = \EX_{x \sim  p_{data}}
  [\log D(x)] + \EX_{z \sim p(z)}[\log(1 - D(G(z)))]
\end{equation}

We modify the vanilla GAN objective by introducing an additional 
discriminator which forces the generator to include features which contribute 
more to the object of interest than its background. P2P uses an adversarial objective using two discriminators, $D_1$ and $D_2$.
The first discriminator $D_1$, discriminates between the real input 
$I_{R}$ and a fake variant of the input, $I_{F_1}$, which consists of the object of 
interest superimposed to a spatially similar training sample, $I_{ctx}$. 
The second discriminator $D_2$ uses a similar approach, but with a 
different fake image, $I_{F_2}$. For this discriminator, the negative area 
around the object of interest is used to create the fake image. 

Fake images used in our GAN formulation are generated by combining the predicted
segmentation mask $\hat{y}$ with our original image $I_R$ and context image $I_{ctx}$
through a process equivalent to alpha blending \citep{porter:alphablend1984}. 

Formally, the process to generate a fake sample is 

\begin{equation}
  f(I, I_{ctx}, y) = y \odot I + y^c \odot I_{ctx}
\end{equation}

where $I, I_{ctx} \in T$, $\odot$ is the Hadamard product and $y$ is a 
segmentation mask. Both $I$ and $I_{ctx}$ are samples from the training dataset $T$, where 
$I_{ctx}$ is found by iterating through adjacent tiles to $I$ until a tile 
without a training label appears, and is chosen to be $I_{ctx}$. Figure 
\ref{fig:contextop} describes the procedure to find $I_{ctx}$ in more detail. 

In P2P's case, $I \equiv I_R$ is the "real" input image fed into the discriminators 
and $\hat{y}$ is used in place of $y$ for generating $I_{F_1}$, while $\hat{y}^c = (1 - \hat{y})$ is used 
for generating $I_{F_2}$. Therefore, the fake images are generated using

\begin{align}
  I_{F_1}  &= f(I_R, I_{ctx}, \hat{y}) = \hat{y} \odot I_R + \hat{y}^c \odot I_{ctx}\\
  I_{F_2}  &= f(I_R, I_{ctx}, \hat{y}^c) = \hat{y}^c \odot I_R + \hat{y} \odot I_{ctx}
\end{align}

The operation $f(\cdot)$ used to generate the fake examples can be thought of as a superimpose operation 
that uses the predicted segmentation mask to crop the object of interest (or its background)
and paste it in a nearby training sample.

Given the above definitions for our real and fake images for $D_1$ and $D_2$, we arrive at the 
following objective:

\begin{align}
  \min_G\max_D\loss(G, D) &= \min_G\max_D \EX_{\chi \sim p_{data}}
  (2 \log(1 - D_1(G_1)) + 2 \log(1 - D_2(G_2)) \nonumber \\ 
    &+ \log(D_1(I_R)) + \log(D_2(I_{ctx})) + \loss_{loc} ) \label{eq:ganobjective}
\end{align}

where $\chi = (I_R, I_{ctx})$, $D = (D_1, D_2)$ and $G = (G_1, G_2)$, $G_1 = f(\hat{y},I_R, I_{ctx})$, 
and $G_2 =  f(\hat{y}^c,I_R, I_{ctx})$. Note that $G_1$ and $G_2$ share weights;
they are both functions of the segmentation model $S(I_R, \tilde{y}) = \hat{y}$, 
although $G_2$ uses the complement of $\hat{y}$, namely $\hat{y}^c$, to generate fake samples. 

The loss function $\loss(G,D)$ for P2P is the sum of $D_1$ and $D_2$'s binary cross entropy loss, 
a generator loss $\loss_G$, and a localization loss $\loss_{loc} = \min(\tilde{y}\cdot \log(\hat{y}) + (1 - \tilde{y})
\cdot \log(1-\hat{y}), \rho)$ that is included in $\loss_G$. 
The localization loss term is a thresholded version of binary cross entropy, 
where $\rho$ is the threshold parameter that controls the amount of 
influence that $\loss_{loc}$ has on the overall loss function. This parameter is chosen with 
a hyperparameter search, specifically ASHA \citep{li:asha2018}. For more details on how the 
objective function is derived, please see the Appendix.

\subsection{Contexts} \label{subsection:context}

\begin{figure}[t!] 
  \centering
  \begin{tikzpicture}[scale=0.65]

    \begin{axis}[hide axis, xmin=0.0, xmax=7.0, ymin=0.0, ymax=7.0]
    
    \draw[help lines] (axis cs:0, 0) grid [step = 1] (axis cs:7,7);
    
    \draw[cyan!65, thick] (3, 6) grid [step=1] (5, 7); 
    \draw[cyan!65, thick] (1, 4) grid [step=1] (3, 5); 
    \draw[cyan!65, thick] (2, 3) rectangle (3, 4); 
    \draw[cyan!65, thick] (5, 3) rectangle (6, 4); 
    \draw[cyan!65, thick] (1, 2) rectangle (2, 3); 
    \draw[cyan!65, thick] (4, 2) grid [step=1] (6, 3); 

    \draw[orange, thick] (1, 6) grid [step=1] (3, 7); 
    \draw[orange, thick] (5, 6) rectangle (6, 7);  
    \draw[orange, thick] (1, 5) grid [step=1] (6, 6); 
    \draw[orange, thick] (4, 4) grid [step=1] (6, 5); 
    \draw[orange, thick] (1, 3) rectangle (2, 4); 
    \draw[orange, thick] (3, 3) grid [step=1] (5, 4); 
    \draw[orange, thick] (2, 2) grid [step=1] (4,3); 

    \draw[violet!85, thick] (3, 4) rectangle (4, 5);

    \draw [fill=orange!25, fill opacity=0.4, draw=none, thick] (1,6) rectangle (3,7); 
    \draw [fill=orange!25, fill opacity=0.4, draw=none, thick] (5,6) rectangle (6,7); 
    \draw [fill=orange!25, fill opacity=0.4, draw=none, thick] (1,5) rectangle (6,6); 
    \draw [fill=orange!25, fill opacity=0.4, draw=none, thick] (4,4) rectangle (6,5); 
    \draw [fill=orange!25, fill opacity=0.4, draw=none, thick] (1,3) rectangle (2,4); 
    \draw [fill=orange!25, fill opacity=0.4, draw=none, thick] (3,3) rectangle (5,4); 
    \draw [fill=orange!25, fill opacity=0.4, draw=none, thick] (2,2) rectangle (4,3); 
  
    \draw [fill=cyan!25, fill opacity=0.6, draw=none, thick] (3,6) rectangle (5,7); 
    \draw [fill=cyan!25, fill opacity=0.6, draw=none, thick] (1,4) rectangle (3,5); 
    \draw [fill=cyan!25, fill opacity=0.6, draw=none, thick] (2,3) rectangle (3,4); 
    \draw [fill=cyan!25, fill opacity=0.6, draw=none, thick] (5,3) rectangle (6,4); 
    \draw [fill=cyan!25, fill opacity=0.6, draw=none, thick] (1,2) rectangle (2,3); 
    \draw [fill=cyan!25, fill opacity=0.6, draw=none, thick] (4,2) rectangle (6,3); 

    \draw [fill=violet!65, fill opacity=0.4, draw=none, thick] (axis cs:3,4) rectangle (axis cs:4,5);
    
    \draw[thick, white] (0,0) rectangle (7, 7);
    \draw[thick, white] (1,2) rectangle (6, 7);
    \draw[ultra thick, white] (2,3) rectangle (5, 6);

    \draw (0,0) rectangle (7, 7); 
    \draw[thick] (1, 2) rectangle (6, 7); 
    \draw[thick, dashed] (axis cs:2, 3) rectangle (axis cs:5, 6); 

    \addplot[only marks, thick, mark=*, mark size =5.5pt, color = gray, fill = gray!20] table[x=x,y=y,col sep=comma]{data/context_data.csv};
    \end{axis}
  \end{tikzpicture}

  \caption{Contexts (blue tiles) are discovered by expanding out from an origin tile, denoted in violet. 
  Origin tiles are tiles containing positive samples (orange) which are indexed during preprocessing 
  then sampled at random during training. We first index the surrounding 8 tiles from the origin. 
  Positive tiles $I_R$ are denoted in orange while contexts $I_{ctx}$ (negative tiles) are denoted in blue. 
  If the number of contexts do not meet the minimum context number, we continuously 
  expand the search space until a sufficient number of positive tiles are met.}
  \label{fig:contextop}
\end{figure}
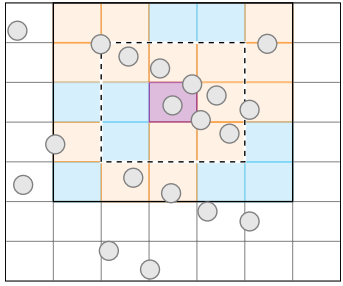

To support the generator in producing realistic fake outputs without being 
overpowered by the discriminator, we need to place these outputs in contexts that 
are realistic and do not expose information which might signal to the discriminator 
that these outputs are fake. As an example we do not want to superimpose objects against a purely black or noisy background. However we do want to place 
these buildings in a suburban neighborhood with trees and roads. We 
define the term context as an image tile in close proximity to a tile in question 
which also does not contain a positive sample. 

P2P requires each context to be extracted from a positive tile’s spatial neighborhood 
to maintain semantic similarity with the rest of the inputs. We provide users the 
flexibility to choose the number of contexts that are aggregated into each batch at 
training time. This ensures a diverse selection of background characteristics for the 
superimpose function. To facilitate dynamic tile fetching during training without 
heavy computational overheads, we map each positive tile to its neighboring contexts 
during a preprocessing step (see Figure \ref{fig:contextop}). The output of this step is a dictionary 
that maps each positive tile to a set number of neighboring contexts which is then 
used by P2P to quickly access these tiles for training.

\subsection{Object Localization}

Since the discriminators do not actually care about whether or not the extracted 
features exhibit sufficient localization behavior, only that these outputs look 
“realistic”, we reinforce the generator with a localization objective. This 
objective makes use of the input point labels, which we term pseudo labels, and creates a buffer around these 
points against which the output segmentation masks are evaluated. In particular, 
we try to minimize the ratio between the amount of positive pixels in the output 
segmentation mask and this buffer. We also set a minimum threshold to how low 
this ratio can go to prevent the model from overweighting the localization objective. 

\section{Implementation}

\begin{figure}[!t]
  \centering
  \begin{tikzpicture}[node distance=1cm]
    \node (f1) {$f_1:$};    
    \node (y_hat) [label=below:$\hat{y}$, right of=f1] {\includegraphics[width=.1\textwidth]{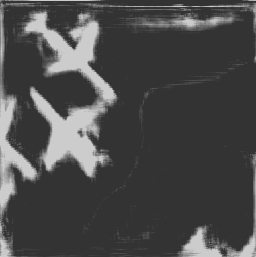}};
    \node (mult) [right of=y_hat]{$\odot$};
    \node (I_R) [label=below:$I_R$, right of=mult]  {\includegraphics[width=.1\textwidth]{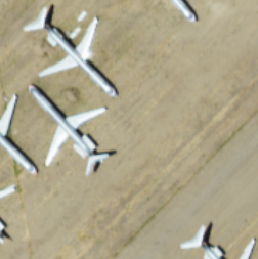}};
    \node (plus) [right of=I_R] {$+$};
    \node (y_hat_c) [label=below:$\hat{y}^c$, right of=plus] {\includegraphics[width=.1\textwidth]{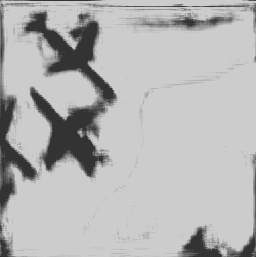}};
    \node (mult2) [right of=y_hat_c]{$\odot$};
    \node (I_ctx) [label=below:$I_{ctx}$, right of=mult2] {\includegraphics[width=.1\textwidth]{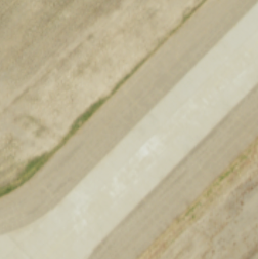}};
    \node (equals) [right of=I_ctx] {$=$};
    \node (I_F1) [label=below:$I_{F_1}$, right of=equals] {\includegraphics[width=.1\textwidth]{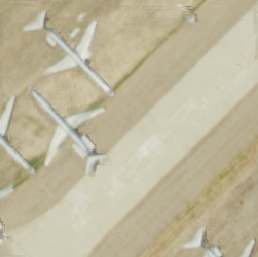}};
  \end{tikzpicture}
    \\
  \begin{tikzpicture}
  \node (f2) {$f_2:$};
  \node (y_hat_prime) [label=below:$\hat{y}^c$, right of=f2] {\includegraphics[width=.1\textwidth]{figs/superimpose/y_hat_c.png}};
  \node (mult) [right of=y_hat_prime]{$\odot$};
  \node (I_R) [label=below:$I_R$, right of=mult] {\includegraphics[width=.1\textwidth]{figs/superimpose/Ir.PNG}};
  \node (plus) [right of=I_R] {$+$};
  \node (y_hat) [label=below:$\hat{y}$, right of=plus] {\includegraphics[width=.1\textwidth]{figs/superimpose/y_hat.PNG}};
  \node (mult2) [right of=y_hat]{$\odot$};
  \node (I_ctx) [label=below:$I_{ctx}$, right of=mult2] {\includegraphics[width=.1\textwidth]{figs/superimpose/Ictx.PNG}};
  \node (equals) [right of=I_ctx] {$=$};
  \node (I_F2) [label=below:$I_{F_2}$, right of=equals] {\includegraphics[width=.1\textwidth]{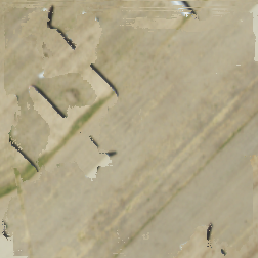}};
  \end{tikzpicture}
  \caption{The general procedure for generating fake images from an input sample $I_R$ and 
  predicted segmentation mask $\hat{y}$. The GAN training procedure is designed to generate segmentation 
  masks $\hat{y}$ that create believable ``fake'' examples.}
\end{figure}

In preprocessing, we divide up the input rasters into multiple image chips. For 
each positive image chip, we search its 8 neighboring contexts in order to index 
any negative chips (chips which do not contain positive samples). If these cannot 
be found in the immediate contexts, we expand our search to the next 16 contexts 
and the next 32 contexts etc. For every positive chip we store a number of its 
negative neighbors (up to $k$ neighbors) in the context dictionary. 

In the training step, we jointly train the segmentation model (generator) 
along with two discriminators. 
We sample a batch of positive chips and feed them to the generator to produce 
output masks. These are then used to ``transplant'' the positive features from 
the original positive context into neighboring negative contexts to produce 
fake positives. Both the real and fake positives are fed into $D_1$ to evaluate 
the positive transplants. 

To prevent the generator from overpowering the discriminator using salient 
but ill-segmented features, we also produce negative contexts. The real 
negative contexts are the original neighbors while the fake negative contexts 
are created by transplanting features from the positive context to its 
neighboring negatives using the inverse of the output mask. These are fed 
to $D_2$. 

Finally at inference time, we simply extract the raw outputs of the generator. 

\section{Experiments}

\begin{table}[b!]
  \caption{Fully-supervised benchmark comparison. }
  \label{benchmark_comparisons}

  \centering
  \resizebox{\textwidth}{!}{
  \begin{tabular}{l|llll|llll|llll|llll|} 
    \toprule
      & \multicolumn{4}{c}{Dice} & \multicolumn{4}{c}{Jaccard} & \multicolumn{4}{c}{Recall} &  \multicolumn{4}{c}{Precision}\\
      \midrule
    Model & \rotate{Well Pads} & \rotate{Airplanes} & \rotate{Woolsey} & \rotate{SpaceNet} 
    & \rotate{Well Pads} & \rotate{Airplanes}  & \rotate{Woolsey} & \rotate{SpaceNet} 
    & \rotate{Well Pads} & \rotate{Airplanes} & \rotate{Woolsey} & \rotate{SpaceNet}
    & \rotate{Well Pads} & \rotate{Airplanes} & \rotate{Woolsey} & \rotate{SpaceNet} \\
    \midrule
    UNet-50 & $0.33$ & $0.65$ & $0.46$ & $0.47$ 
    & $0.20$  & $0.49$ & $0.31$  & $0.31$ 
    & $\textbf{1.00}$ & $\textbf{1.00}$ & $\textbf{0.99}$ & $\textbf{1.00}$ 
    & $0.20$  & $0.49$ & $0.31$ & $0.31$\\
    UNet-101 & $0.21$ & $0.46$& $0.57$ & $0.47$ & $0.12$  & $0.30$ & $0.41$ 
    & $0.32$  & $\textbf{1.00}$ & $\textbf{1.00}$ & $0.85$ & $\textbf{1.00}$ & $0.12$ & $0.30$ & $0.44$ & $0.32$\\
    UNet-50 (CE) & $\textbf{0.80}$ & $\textbf{0.90}$ & $\textbf{0.81}$ & $\textbf{0.90}$ 
    & $\textbf{0.68}$  & $\textbf{0.82}$ & $\textbf{0.70}$ & $\textbf{0.82}$ 
    & $0.90$ & $0.86$ & $0.80$ & $0.89$ 
    & $\textbf{0.75}$ & $\textbf{0.94}$ & $\textbf{0.85}$ & $\textbf{0.92}$\\
    P2P (ours) & $0.65$ & $0.64$ & $0.66$ & $0.62$ & $0.49$  
    & $0.48$ & $0.51$ &  $0.46$ & $0.93$ & $0.67$ & $0.74$ & $0.60$ & $0.52$ & $0.64$ & $0.64$ & $0.68$  \\

    \bottomrule
  \end{tabular} }
\end{table}

\subsection{Datasets}

In our experiments, we make use of five datasets to train P2P and assess its performance. 
This includes two small datasets for well pads and airplanes as described in the Appendix, 
and one medium dataset for building footprint extraction which we 
sourced from USAA. We also train and benchmark our model against the SpaceNet Challenge 
AOI2 Las Vegas dataset \citep{vanetten:spacenet2018}. Additionally, as part of our ablative studies we also 
create a smaller version of this dataset which we name Vegas Lite. 

\paragraph{Well pad dataset}
We construct a well pad dataset using NAIP: Natural Color imagery at 1.5m resolution. 
This dataset is created using ArcGIS Pro and contains 9 training rasters and 1 testing 
raster. There are 2909 training samples and 426 testing samples, which makes this a 
small-to-medium sized dataset. The labels are bounding geometries of visible well pads, 
which contain both active and disused examples. 

\paragraph{Airplane dataset}
We construct an airplane dataset using NAIP: Natural Color imagery at 0.5m resolution. 
This dataset is created using ArcGIS Pro and contains 2 training rasters and 1 testing 
raster. Given there are 335 training samples and 40 testing samples, this constitutes 
a small dataset and what can be realistically expected in a real-world proof-of-concept 
application. The labels incorporate a variety of plane sizes from 3 locations in the 
United States: Southern California Logistics Airport, Roswell International Air 
Center and Phoenix Goodyear Airport. 

\paragraph{Woolsey Fire dataset}

We make use of the DataWing imagery from USAA that was captured after the 2018 
Woolsey Fires for damage assessment. These are 0.3m resolution rasters with 
8682 training samples distributed across 6 training rasters and 572 samples on 
1 testing raster. This is a medium sized dataset and contains a large number of 
negative samples (of damaged houses) in a variety of terrains. 

\paragraph{SpaceNet Challenge: Las Vegas dataset}
In order to benchmark our model against popular datasets for remote sensing, 
we make use of the SpaceNet Las Vegas (AOI2) dataset \citep{vanetten:spacenet2018}. Both training and 
testing sets are publicly available however the testing set does 
not contain ground truth geometries. We therefore further split the training 
set into 2905 image tiles for training (tiles numbered 1325 and above) and 712 image 
tiles for testing (tiles up to, but not including, 1325).

\paragraph{Vegas Lite dataset}

We further reduce the size of the Las Vegas dataset in order to perform ablative 
tests on a smaller set of samples. This we call the Vegas Lite dataset and is simply 
the full Las Vegas Dataset without any tiles numbered 3000 upwards in the training 
set and 500 upwards in the testing set. This dataset contains 970 tiles in the 
training set and 268 tiles in the testing set. These are roughly 33\% and 38\% 
the size of the original training and testing sets respectively, assuming samples 
are equally distributed across tiles. 

We use the Vegas Lite dataset to understand the importance of contextual similarity 
between positive and negative chips. We do so by introducing a series of negative 
rasters with varying contextual similarities in our ablative tests.

\subsection{Performance Study}
We evaluate our model based on several real-world feature segmentation tasks 
in order show the efficacy of our proposed approach. 

\paragraph{P2P vs Baseline Comparison}

We compare P2P which uses weak labels $\tilde{y}$ against 3 fully supervised segmentation 
models which make use of ground truth polygon labels $y$. Among baselines trained with Dice loss, we find 
that UNet-50 produces the best results on all except the well pads dataset, 
where UNet-18 performs the best out of the baseline models. In the case of well pad extraction, an increase in the 
number of parameters in the model results in lower performance, suggesting that 
a low-complexity problem such as well pad extraction could be over-parameterized in this
approach. A considerable improvement is made to the performance of baseline models with full labels 
when trained with cross-entropy loss, which produces better learnable gradients for these datasets.   
A survey of the visual outputs show that these baselines generally produce segmentation 
masks that are significantly larger than the object outlines which is not the 
case for P2P. We further clarify that these P2P results consider the optimal 
localization guidance parameters, however it is not difficult to select 
a uniform guidance parameter that would outperform all baselines (7000 for 
example). 

\subsection{Ablative Tests}

\paragraph{Negative Discriminator} 

We explore the impact of a second negative discriminator on testing accuracy 
through all 4 datasets. We define this second discriminator $D_2$ as a UNet-50 (same 
as $D_1$) binary classifier which predicts the probability that its input comes 
from the distribution of real negatives. Negative samples associated with a 
specific positive sample are drawn from the positive sample’s neighboring contexts. 
This is done during preprocessing as discussed in Section 3.4. We then employ the context 
shift function $f$ to superimpose the positive sample $I_R$
to its context $I_{ctx}$ using the inverse of the labels. This creates our fake negative sample $I_{F_2}$, 
which we jointly feed to $D_2$ along with the real context $I_{ctx}$.

In experimentation, we find that implementing $D_2$ significantly improves model 
performance across all but one of the datasets. In the well pads 
dataset, we find that introducing $D_2$ incurs a small performance cost. This is 
possibly due to the simple nature of the segmentation task where having additional 
tunable parameters potentially lead to model overfitting.  

By observing the quantitative and visual outputs we identify that the lack of $D_2$ 
generally leads to segmentation masks with much lower precision and higher recall.

\begin{table}[h!]
  \caption{Negative Discriminator $D_2$ ablative study.}
  \label{table:d2_ablative}

  \centering
  \resizebox{\textwidth}{!}{
  \begin{tabular}{l|lll|lll|lll|lll|} 
    \toprule
      & \multicolumn{3}{c}{Dice} & \multicolumn{3}{c}{Jaccard} & \multicolumn{3}{c}{Recall} &  \multicolumn{3}{c}{Precision}\\
      \midrule
    Model & \rotate{Well Pads} & \rotate{Airplanes} & \rotate{Woolsey}  
    & \rotate{Well Pads} & \rotate{Airplanes}  & \rotate{Woolsey} 
    & \rotate{Well Pads} & \rotate{Airplanes} & \rotate{Woolsey}
    & \rotate{Well Pads} & \rotate{Airplanes} & \rotate{Woolsey} \\
    \midrule
    P2P w/o $D_2$ 
    & $\textbf{0.67}$ & $0.23$& $0.43$ 
    & $\textbf{0.51}$ & $0.13$ & $0.28$  
    & $0.92$ & $\textbf{0.96}$ & $\textbf{0.90}$ 
    & $\textbf{0.54}$ & $0.13$ & $0.30$  \\
    P2P w/ $D_2$ 
    & $0.65$ & $\textbf{0.64}$ & $\textbf{0.66}$ 
    & $0.49$  & $\textbf{0.48}$ & $\textbf{0.51}$
    & $\textbf{0.93}$ & $0.67$ & $0.74$ 

    & $0.52$ & $\textbf{0.64}$ & $\textbf{0.64}$  \\

    \bottomrule
  \end{tabular} }
\end{table}

\paragraph{Localization Guidance}

We show that a purely unsupervised approach without any guidance on object localization 
does not produce good results in practice. In order to guide the 
model during training, we create a buffer around object centroids at training time 
and incur a penalty on the generator whenever predicted masks do not cover these
buffered pixels. We then threshold this loss to prevent this localization loss 
from overwhelming the generator. A failure case would be the model treating the 
buffered centroids as the ground truth labels under a fully supervised objective. 
By thresholding, we ensure that the model is only concerned with localization at 
a rough scale. 

To qualify the optimal buffer size for a variety of datasets we trained P2P using 
a variety of preset buffer sizes. This is a variable we term centroid 
size multiplier (csm). The exact formulation for localization loss and additional visual results can be found 
in the Appendix.

We also recognize that a shortfall of this method is that it is less suitable for 
detecting objects that do not share the same size. This is of lesser concern for 
remote sensing use cases but should be taken into consideration when applying this 
model to general-purpose segmentation tasks. 

\begin{table}[h!]
  \caption{Localization Guidance ablative study for various $\textmd{csm}$ values.}
  \label{table:localization_ablative}

  \centering
  \resizebox{\textwidth}{!}{
  \begin{tabular}{l|lll|lll|lll|lll|} 
    \toprule
      & \multicolumn{3}{c}{Dice} & \multicolumn{3}{c}{Jaccard} & \multicolumn{3}{c}{Recall} &  \multicolumn{3}{c}{Precision}\\
      \midrule
    Model (csm) & \rotate{Well Pads} & \rotate{Airplanes} & \rotate{Woolsey} 
    & \rotate{Well Pads} & \rotate{Airplanes}  & \rotate{Woolsey} 
    & \rotate{Well Pads} & \rotate{Airplanes} & \rotate{Woolsey} 
    & \rotate{Well Pads} & \rotate{Airplanes} & \rotate{Woolsey}  \\
    \midrule
    P2P ($6000$) & $0.27$ & $0.15$& $0.04$ 
    & $0.16$ & $0.09$ & $0.02$ 
    & $0.72$ & $0.17$ & $0.02$ 
    & $0.17$ & $0.18$ & $0.17$ \\
    P2P ($7000$) & $0.55$ & $\textbf{0.64}$ & $\textbf{0.66}$ 
    & $0.39$  & $\textbf{0.48}$ & $\textbf{0.51}$ 
    & $\textbf{0.87}$ & $0.67$ & $0.74$ 
    & $0.42$ & $0.64$ & $0.64$  \\
    P2P ($8000$) & $0.64$ & $0.54$ & $0.54$
    & $0.48$  & $0.38$ & $0.38$ 
    & $0.84$ & $0.78$ & $0.49$ 
    & $0.54$ & $0.42$ & $0.64$ \\
    P2P ($9000$) & $0.07$ & $0.56$ & $0.64$
    & $\textbf{0.50}$  & $0.39$ & $0.48$ 
    & $0.78$ & $\textbf{0.86}$ & $\textbf{0.82}$ 
    & $0.60$ & $0.43$ & $0.55$ \\
    P2P ($10000$) & $\textbf{0.65}$ & $0.51$ & $0.65$
    & $0.49$  & $0.35$ & $0.50$ 
    & $0.52$ & $0.37$ & $0.62$ 
    & $\textbf{0.93}$ & $\textbf{0.86}$ & $\textbf{0.74}$  \\

    \bottomrule
  \end{tabular} }
\end{table}

\paragraph{Contextual Similarity}\label{subsection:contextual_ablative}
In order to understand the importance of applying semantically coherent
negative contexts to $f$, we try out different types of context images in 
training P2P on the Vegas Lite dataset. We compare the performance of the 
Original Context image (an image that is spatially similar to real image 
tiles) with transformed context rasters that consist of zero values (Blank 
Context), chromatic aberrations (Red Context) and random rasters drawn from 
a Gaussian distribution (Noise Context). We note that that the introduction 
of strong signals indicative of contexts such as in the case of Red Context 
results in poor model performance whereas spatially similar contexts produce 
the highest Dice/Jaccard scores. One outlier example is the Noise Context, 
which produces equivalently good Dice/Jaccard scores as the Original but instead 
has much lower precision and higher recall. This is indicative of high model bias. 
We further highlight the similarity between this and removing the contextual discriminator 
which also results in lower precision and higher recall. Noisy contexts can therefore be 
interpreted as effectively weakening the learning signal from the contextual discriminator.

\begin{table}[h!]
  \caption{Contextual similarity ablative study using only the Vegas Lite dataset.}
  \label{table:context_ablative}
  \centering
  \begin{tabular}{l|l|l|l|l} 
    \toprule
    Transformation & Dice & Precision & Recall & Jaccard\\
    \midrule
    Original Context & $\textbf{0.66}$ & $\textbf{0.79}$ & $0.58$ & $\textbf{0.50}$ \\
    Blank Context & $0.60$ & $0.56$ & $0.70$ & $0.44$ \\
    Red Context & $0.56$ & $0.57$ & $0.61$ & $0.40$ \\
    Noise Context & $\textbf{0.66}$ & $0.57$ & $\textbf{0.84}$ & $\textbf{0.50}$ \\
    \bottomrule
  \end{tabular} 
\end{table}

\section{Conclusion \& Future Work}

In this paper, we introduced Points2Polygons, a semantic segmentation model that uses an adversarial 
approach consisting of two discriminators to learn segmentation masks using weak labels. 
We show that a generative adversarial network can be used to perform segmentation 
tasks with good performance across a wide variety of small datasets where only weak 
point labels are provided by introducing the contextual superimpose function. 

Future work includes evaluating our weak label approach to other forms of imagery 
outside of remote sensing. The remote sensing perspective on this problem is important 
since spatially neighboring
tiles can be used to construct new contexts to improve the models robustness, but perhaps the same
technique could be applied in different settings. While this approach is powerful by itself to produce 
segmentation using simple labels, it would be more powerful when combined in a object detection setting
to provide instance segmentation with weak labels.

\begin{ack}

  We would like to thank our colleague Mansour Raad for his domain expertise that made this research possible, as well as 
  Omar Maher, Ashley Du and Shairoz Sohail for reviewing and providing comments on a draft of this paper. 
\end{ack}

\bibliography{references}

\newpage
\appendix

\section{Appendix}

\subsection{Derivation of the modified GAN objective function}

Points2Polygons uses an objective function similar to that of a traditional generative adversarial network (GAN). 
In this section we show the 
connection between the traditional objective of a GAN with discriminator $D$ and generator $G$ and 
our modified GAN with discriminators $D_1$ and $D_2$ and generator $G$. Recall from Section \ref{subsection:adversarial} 
that our generator differs from a typical generator in that it is not a function of a latent variable $z$. Rather, 
it is a function of our segmentation mask $\hat{y}$, input image $I_R$, and context image $I_{ctx}$, where both 
$I_R$ and $I_{ctx}$ come from our training data $T$. We also show that our generator is an adaptive superimpose 
function that improves P2P's segmentation masks by utilizing information from an image's neighboring contexts.

The first difference between the traditional GAN formulation and ours is that rather than maximizing the loss of a 
single discriminator $D$, we 
maximize the loss of two discriminators, $D_1$ and $D_2$, with respect to their parameters. Given a real image
$I_R$ and a generated positive image $I_{F_1}$, we have the following loss for $D_1$:

\begin{align}
  \loss_{D_1} &= \EX_{I_R, I_{ctx} \sim p_{data}}[\log(1 - D_1(I_{F_1})) + \log(D_1(I_R))].
\end{align}

Some literature minimizes the negative loss, however we use the above form to remain consistent with the 
$\min\max$ formulation introduced in Section \ref{subsection:adversarial}. Hence, when referring to the ``loss''
function of $D_1$ and $D_2$ in this section, it is implied that we are generally referring to a discriminator's cost function.

Note that $I_{F_1} = f(\hat{y}, I_R, I_{ctx})$, where $f$ is the superimpose function covered in Section \ref{subsection:context}.
Additionally, $\hat{y} = S(\tilde{y}, I_R)$ is the output of our segmentation model, which leads us to the following 
form of $\loss_{D_1}$:

\begin{align}
  \loss_{D_1} &= \EX_{I_R, I_{ctx} \sim p_{data}}[\log(1 - D_1(f(\hat{y}, I_R, I_{ctx}))) + \log(D_1(I_R))]\nonumber \\
  &= \EX_{I_R, I_{ctx} \sim p_{data}}[\log(1 - D_1(f(S(\tilde{y}, I_R), I_R, I_{ctx}))) + \log(D_1(I_R))].\label{eq:LD1}
\end{align}

Therefore, the objective of $D_1$ is to generate a set of model parameters that can discriminate between an object in its 
proper context (real) and that same object in a neighboring context (fake). 

The second discriminator loss function $\loss_{D_2}$ follows the same pattern, except our ``real'' image is our neighboring 
context $I_{ctx}$ and our ``fake'' image relies on the complement of the segmentation mask $\hat{y}^c$:

\begin{align}
  \loss_{D_2} &= \EX_{I_R, I_{ctx} \sim p_{data}}[\log(1 - D_2(I_{F_2})) + \log(D_2(I_{ctx}))]\nonumber \\
  &= \EX_{I_R, I_{ctx} \sim p_{data}}[\log(1 - D_2(f(\hat{y}^c, I_R, I_{ctx}))) + \log(D_2(I_{ctx}))]\nonumber \\
  &= \EX_{I_R, I_{ctx} \sim p_{data}}[\log(1 - D_2(f(S(\tilde{y}, I_R)^c, I_R, I_{ctx}))) + \log(D_2(I_{ctx}))].\label{eq:LD2}
\end{align}

Finally, P2P minimizes the generator loss function $L_{G}$ with the goal of generating segmentation masks that 
pastes the object of interest into its neighboring context in a realistic way:

\begin{align}
  \loss_{G} =\EX_{I_R, I_{ctx} \sim p_{data}}&[\log(1 - D_1(I_{F_1})) + \log(1 - D_2(I_{F_2})) + \loss_{loc}]\nonumber \\
  = \EX_{I_R, I_{ctx} \sim p_{data}}&[\log(1 - D_1(f(\hat{y}, I_R, I_{ctx}))) + \log(1 - D_2(f(\hat{y}^c, I_R, I_{ctx}))) + \loss_{loc}]\nonumber \\
  = \EX_{I_R, I_{ctx} \sim p_{data}}&[\log(1 - D_1(f(S(\tilde{y}, I_R), I_R, I_{ctx})))\nonumber \\
  &+ \log(1 - D_2(f(S(\tilde{y}, I_R)^c, I_R, I_{ctx}))) + \loss_{loc}].\label{eq:LG}
\end{align}

where

\begin{equation}
  \loss_{loc} = \min (\tilde{y} \cdot \log(\hat{y}) + (1 - \tilde{y}) \cdot \log(1-\hat{y}), \rho)
\end{equation}

is a thresholded localization loss controlled by a scalar threshold parameter $\rho$. 

Given Equations \ref{eq:LD1}, \ref{eq:LD2} and \ref{eq:LG}, it is clear that the generator for 
our GAN is either $f(S(\tilde{y}, I_R), I_R, I_{ctx})$ or $f(S(\tilde{y}, I_R)^c, I_R, I_{ctx})$ for
$D_1$ or $D_2$, respectively. 

 Therefore, we use the following notation for the rest of the derivation:

\begin{align}
  G_1 &= f(S(\tilde{y}, I_R), I_R, I_{ctx})\\
  G_2 &= f(S(\tilde{y}, I_R)^c, I_R, I_{ctx}).
\end{align}

Since $f$ is a superimpose function that utilizes the ouput from a segmentation 
model $S$, we remark that $G_1$ and $G_2$ are adaptive superimpose functions that are optimized 
to produce high-quality fake examples with each backwards pass of $S$. 

Finally, the losses above are combined in the following way:

\begin{equation}
  \loss(G,D) = \loss_G + \loss_{D_1} + \loss_{D_2}. \label{eq:LGD}
\end{equation}

If we plug in Equations \ref{eq:LD1}, \ref{eq:LD2} and \ref{eq:LG} into Equation \ref{eq:LGD}, 
combine common terms, use the new notation for $G_1$ and $G_2$, and apply our $\min$ and $\max$ operations
to both sides, we arrive at:

\begin{align*}
  \min_G\max_D\loss(G, D) &= \min_G\max_D \EX_{I_R, I_{ctx} \sim p_{data}}
  [2 \log(1 - D_1(G_1)) + 2 \log(1 - D_2(G_2))\\ 
    &+ \log(D_1(I_R)) + \log(D_2(I_{ctx})) + \loss_{loc} ]
\end{align*}

which is our GAN's objective function from Section \ref{subsection:adversarial}.  $\blacksquare$

\begin{figure}[!t]
  \begin{tabular}{ccc}
    \includegraphics[width=0.3\textwidth]{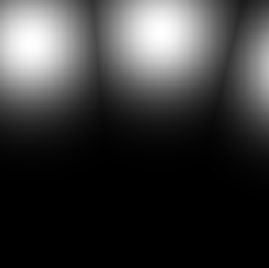} & 
    \includegraphics[width=0.3\textwidth]{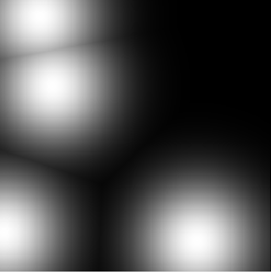} & 
    \includegraphics[width=0.3\textwidth]{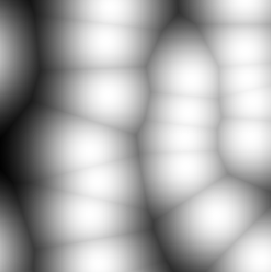} 
  
  \end{tabular}
  \caption{Centroid rasters generated by creating a multivariate Gaussian for every point. The pseudo label $\tilde{y}$ is created 
  from the centroid raster using a threshold constant. Please see Section \ref{app:context_ablative} for more details 
  regarding how the centroid rasters are constructed.}
\end{figure}

\begin{figure}[!t]
  \begin{center}
  \setlength{\tabcolsep}{1pt}
    \begin{tabular}{ccccccc}
  \includegraphics[width=.14\linewidth]{figs/bangers/planes/l23_P2P/Ir_6.png} &
   \includegraphics[width=.14\linewidth]{figs/bangers/planes/l23_P2P/y_tilde_6.png} &
   \includegraphics[width=.14\linewidth]{figs/bangers/planes/l210_UNET101/y_hat_4.png} &
   \includegraphics[width=.14\linewidth]{figs/bangers/planes/l211_UNET50/y_hat_4.png} &
   \includegraphics[width=.14\linewidth]{figs/aircraft_unet50.png} & 
   \includegraphics[width=.14\linewidth]{figs/bangers/planes/l23_P2P/y_hat_7.png} &
   \includegraphics[width=.14\linewidth]{figs/bangers/planes/l23_P2P/y_6.png} \\
   \includegraphics[width=.14\linewidth]{figs/bangers/woolsey/l13_P2P/Ir_2.png} &
   \includegraphics[width=.14\linewidth]{figs/bangers/woolsey/l13_P2P/y_tilde_2.png} &
   \includegraphics[width=.14\linewidth]{figs/bangers/woolsey/l110_UNET18/y_hat_1.png} &
   \includegraphics[width=.14\linewidth]{figs/bangers/woolsey/l111_UNET50/y_hat_1.png} & 
   \includegraphics[width=.14\linewidth]{figs/woolsey_unet50.png} & 
   \includegraphics[width=.14\linewidth]{figs/bangers/woolsey/l13_P2P/y_hat_2.png} &
   \includegraphics[width=.14\linewidth]{figs/bangers/woolsey/l13_P2P/y_2.png} \\
   \includegraphics[width=.14\linewidth]{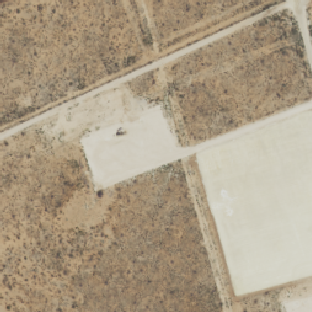} &
   \includegraphics[width=.14\linewidth]{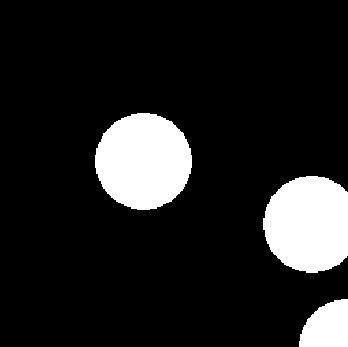} &
   \includegraphics[width=.14\linewidth]{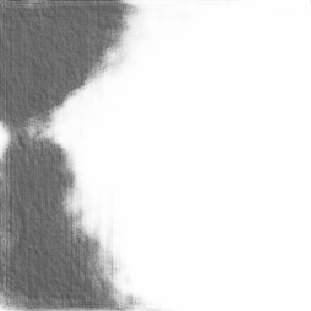} & 
   \includegraphics[width=.14\linewidth]{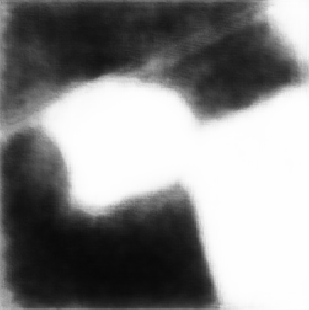} & 
   \includegraphics[width=.14\linewidth]{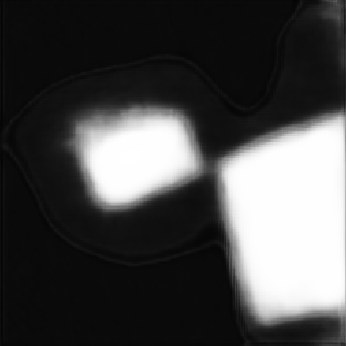} & 
   \includegraphics[width=.14\linewidth]{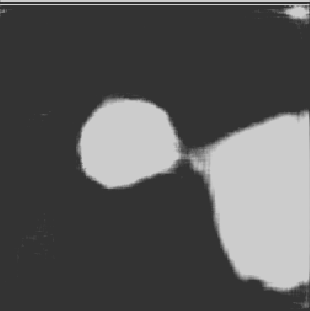} &
   \includegraphics[width=.14\linewidth]{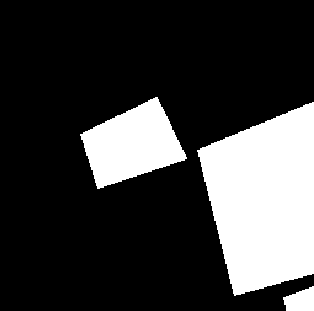} \\
   \includegraphics[width=.14\linewidth]{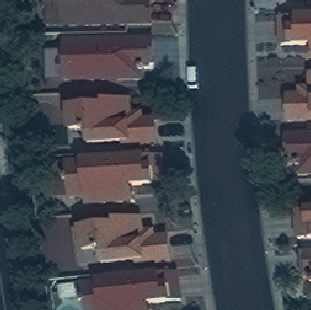} &
   \includegraphics[width=.14\linewidth]{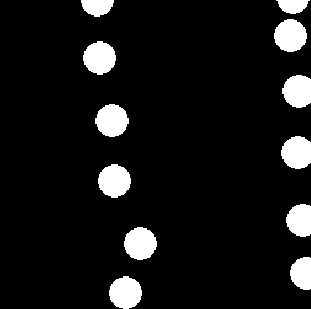} &
   \includegraphics[width=.14\linewidth]{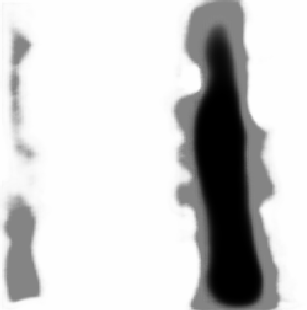} & 
   \includegraphics[width=.14\linewidth]{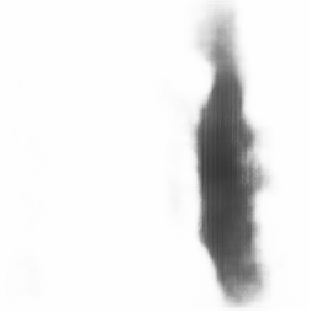} & 
   \includegraphics[width=.14\linewidth]{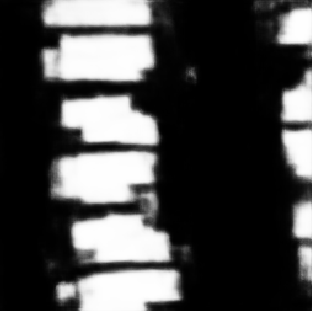} & 
   \includegraphics[width=.14\linewidth]{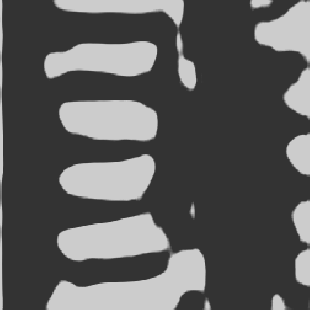} &
   \includegraphics[width=.14\linewidth]{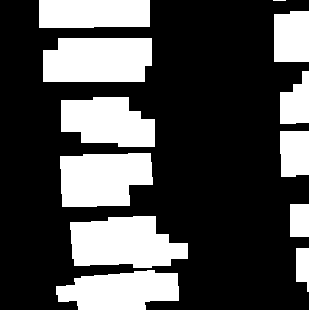} \\
   (a) & (b)  & (c) & (d) & (e) & (f)  & (g)  \\
  \end{tabular}
  
  \end{center}
  \caption[Predicted segmentation masks from the Aircraft and Woolsey datasets]{
    Extended results for predicted segmentation masks from the Aircraft (row 1), Woolsey (row 2), Well Pad (row 3),
    and SpaceNet (row 4) datasets using fully-supervised
    UNet models with ResNet backbones (c-e) and our semi-supervised P2P model (f). All generated masks
    come from training with Dice loss, besides (e), which is trained with cross entropy loss for comparison. 
    (a) Input image to each model. (b) Pseudo label $\tilde{y}$ used to train P2P. (c) Output from 
    a UNet-101 baseline model. (d) Output from UNet-50. (e) Output from UNet-50 trained with cross-entropy. 
    (f) Output from P2P (ours). (g) Ground Truth labels. 
  \label{fig:bangers_p2} }
  
\end{figure}

\subsection{Localization Guidance ablative study visual results}\label{app:context_ablative}
\begin{figure}[!t]
  \begin{center}
  \setlength{\tabcolsep}{1pt}
  \begin{tabular}{cccccccc}
    Input $I_R$ & Context $I_{ctx}$ & Pseudo label $\tilde{y}$ & Output $\hat{y}$ & Pos. Fake $I_{F_1}$ & Neg. Fake $I_{F_2}$ & Ground Truth $y$ \\ 
  \includegraphics[width=0.14\textwidth]{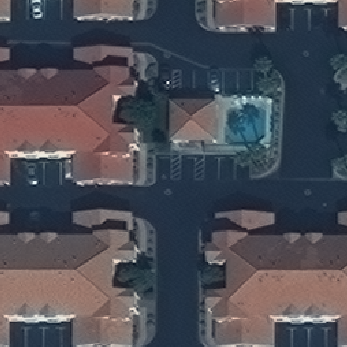} & 
   \includegraphics[width=0.14\textwidth]{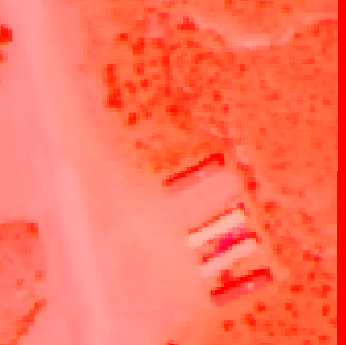} & 
   \includegraphics[width=0.14\textwidth]{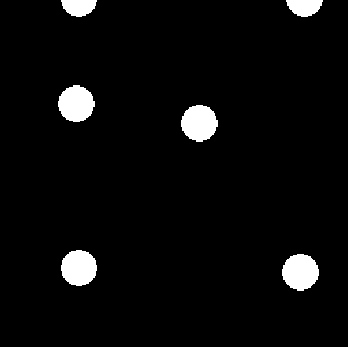} &
   \includegraphics[width=0.14\textwidth]{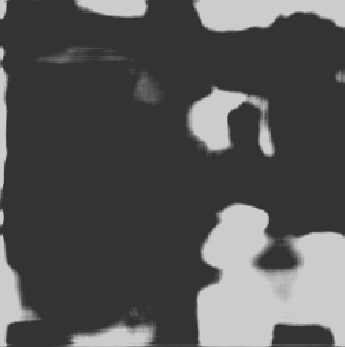} &
   \includegraphics[width=0.14\textwidth]{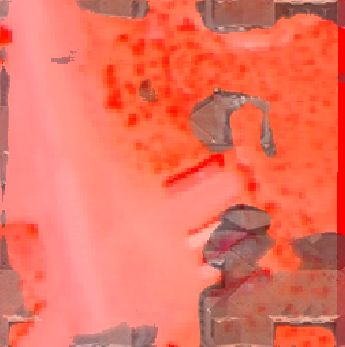} &
   \includegraphics[width=0.14\textwidth]{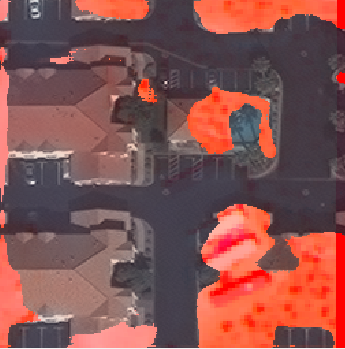} & 
   \includegraphics[width=0.14\textwidth]{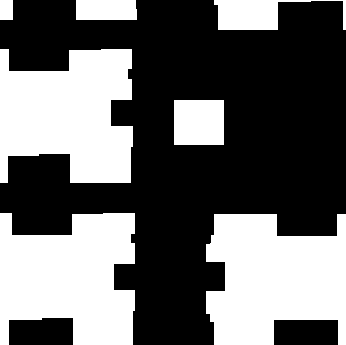} 
   \\ 

   \includegraphics[width=0.14\textwidth]{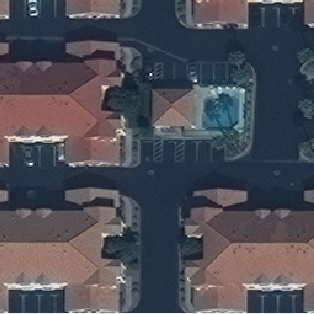} & 
   \includegraphics[width=0.14\textwidth]{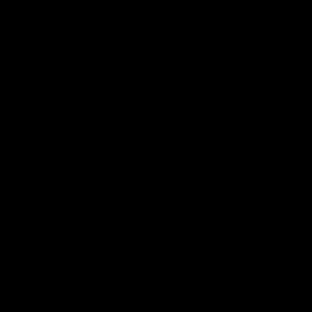} & 
   \includegraphics[width=0.14\textwidth]{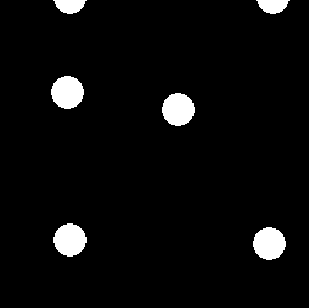} &
   \includegraphics[width=0.14\textwidth]{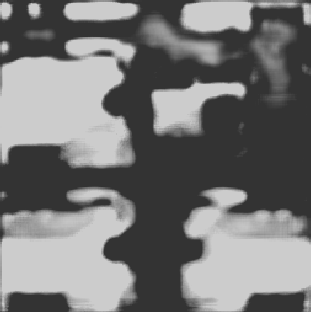} &
   \includegraphics[width=0.14\textwidth]{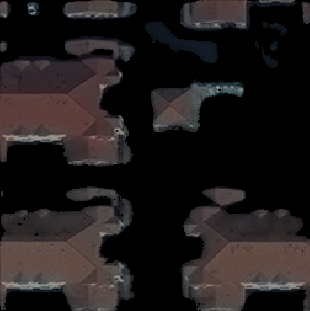} &
   \includegraphics[width=0.14\textwidth]{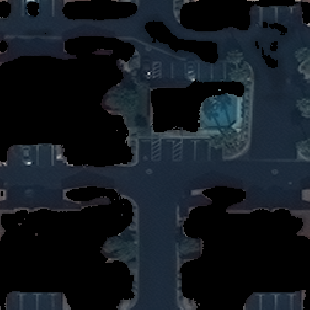} & 
   \includegraphics[width=0.14\textwidth]{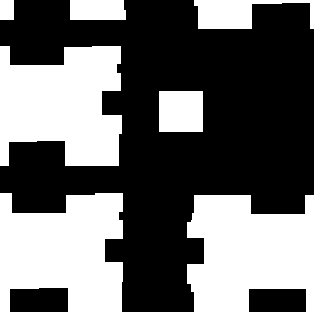} 
   \\ 

   \includegraphics[width=0.14\textwidth]{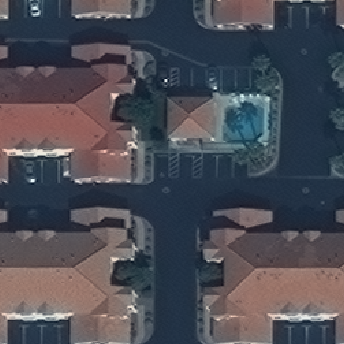} & 
   \includegraphics[width=0.14\textwidth]{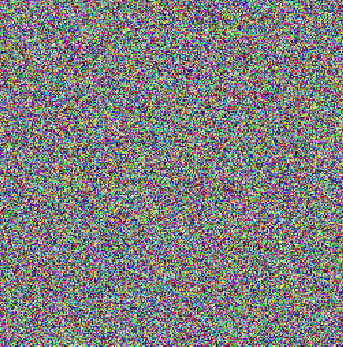} & 
   \includegraphics[width=0.14\textwidth]{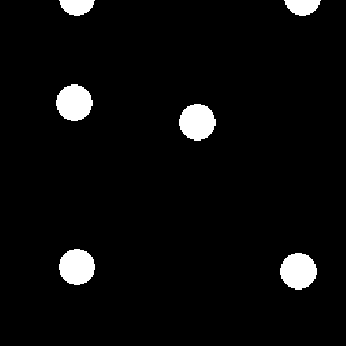} &
   \includegraphics[width=0.14\textwidth]{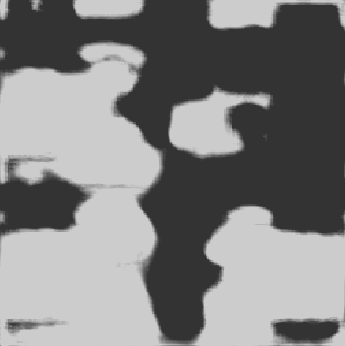} &
   \includegraphics[width=0.14\textwidth]{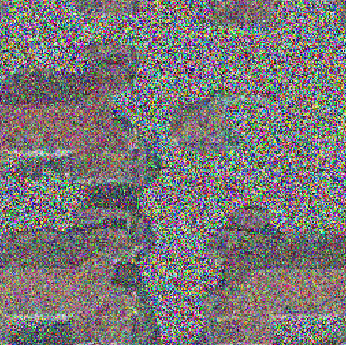} &
   \includegraphics[width=0.14\textwidth]{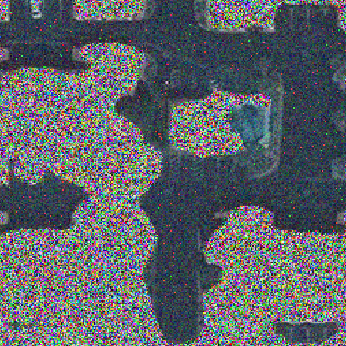} & 
   \includegraphics[width=0.14\textwidth]{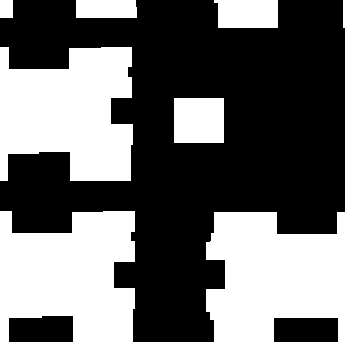} 
   \\ 

   \includegraphics[width=0.14\textwidth]{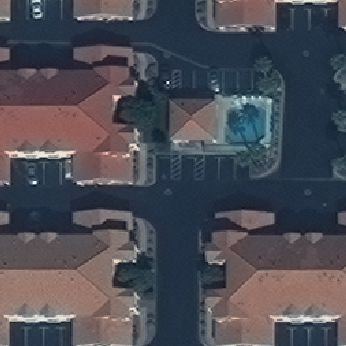} & 
   \includegraphics[width=0.14\textwidth]{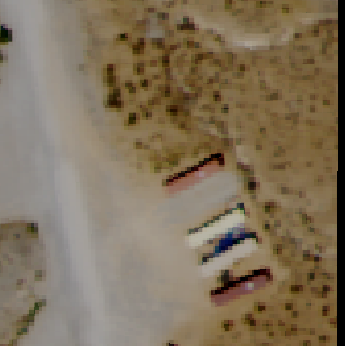} & 
   \includegraphics[width=0.14\textwidth]{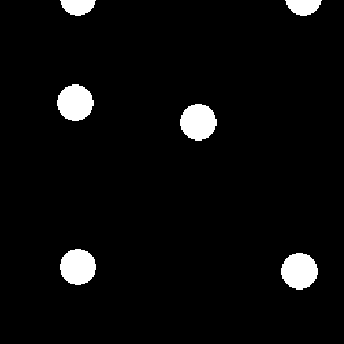} &
   \includegraphics[width=0.14\textwidth]{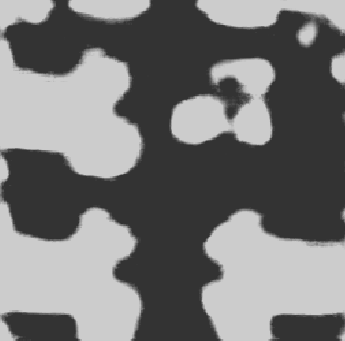} &
   \includegraphics[width=0.14\textwidth]{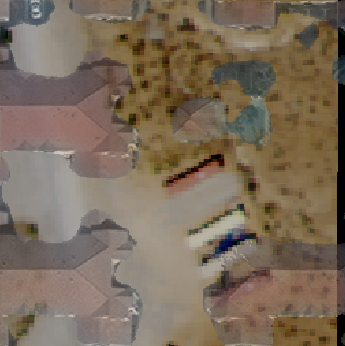} &
   \includegraphics[width=0.14\textwidth]{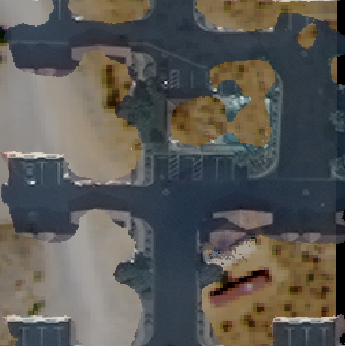} & 
   \includegraphics[width=0.14\textwidth]{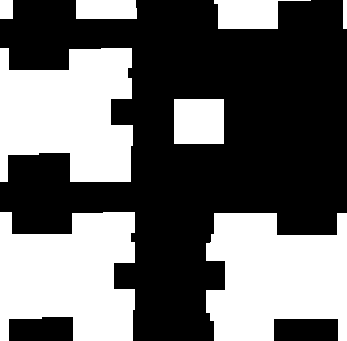} 
   \\ 

  \end{tabular}
  \end{center}
  \caption[Effect of different transformations on the context image and resulting segmentation mask.]{
    Effect of different transformations on the context image and resulting segmentation mask. Please see
    Section \ref{subsection:contextual_ablative} for further details.
  \label{fig:ablative_results}}
\end{figure}

In this ablative study we observe the effect that changing the localization guidance buffer size has on the output prediction $\hat{y}$. 
This buffer size is determined by the centroid size multiplier (csm) variable which is factored into the localization loss for the generator. 
See Algorithm \ref{app:localization_algo} for more details on the role of csm in the buffer size calculation and localization loss.

\begin{algorithm}[!t]\label{app:localization_algo}
  \SetAlgoLined
  \KwIn{constant csm, constant $\Sigma$, constant $\gamma$, constant $\rho$}
  canvas $\leftarrow$ zeros \;
  $p_{geom} \leftarrow \textmd{label}(I_R)$\;
  \For{point $p \in p_{geom}$}{
  $\mu \leftarrow (p_x, p_y)$\;
  $\textmd{centroid\_raster} \sim MVN(\mu, \Sigma)$\;
  $\textmd{canvas} \leftarrow  \textmd{element-wise max}(\textmd{canvas}, \textmd{centroid\_raster})$\;
  
  }
  $\textmd{canvas}^* \leftarrow \textmd{csm} \cdot \textmd{canvas}$\;
  $\tilde{y} \leftarrow \textmd{element-wise where}(\textmd{canvas}^* \geq \gamma, 1, 0)$\;
  $\loss_{loc} \leftarrow \min(\textmd{cross-entropy}(\hat{y}, \tilde{y}), \rho)$\;
  \KwRet{$\loss_{loc}$}
  \caption{Localization loss formulation per image sample $I_R$.}
  \end{algorithm}

In the procedure for deriving $\tilde{y}$, we use a bivariate Gaussian with covariance matrix $\Sigma$ to generate 
a candidate buffer area around each point label $p$, whose coordinates define the mean. A hyperparameter $\gamma$  is used as a centroid threshold constant to binarize the candidate buffer areas.
We employ a threshold parameter $\rho$ to control the influence of the localization loss on the overall generator loss. 
The csm variable then effectively determines the size of the buffered centroids. 

Through experimentation we observe that having a smaller buffer (csm=$7000$) produces a better Dice score and better visual results for most datasets, with the notable exception of the well pads dataset, where having a larger buffer produces a higher Dice score. 
We argue that this is due to the fact that a larger buffer effectively overpowers the other loss terms in the generator and brings in contextual noise in cases where the buffer extends beyond the object’s bounding geometry, as is the case for planes or smaller buildings 
but not for well pads which tend to be larger.

\subsection{Pre/Post Processing}\label{app:postprocessing}

We use the ArcGIS Pro geospatial software in order to perform input preprocessing on all datasets. 
Preprocessing consists of modifying image bit depth (we use uint8 for all datasets) as well as cell size adjustments which define the image resolution. 
We also use ArcGIS Pro to transform the labels (such as for data obtained from the SpaceNet dataset) from .geojson into ESRI .shp files using the \texttt{JSON to Features} geoprocessing (GP) tool. 
Preprocessing further consists of tiling input rasters into smaller, manageable chips using the \texttt{Export Training Data for Deep Learning} GP tool. 
Note the preprocessing steps are not needed when directly using our provided datasets. 

To produce the actual polygonal outputs (i.e. the Polygons in Points2Polygons), we first apply a threshold of $0.5$ to convert the output $\hat{y}$ to a binary representation
before converting the mask to a polygon geometry using the \texttt{Feature to Polygon} GP tool.  For 
certain use cases such as building footprints and well pads we also use the \texttt{Building Footprint Regularization} GP tool to enhance our results.  

\begin{figure}[!t]
  \begin{center}
  \setlength{\tabcolsep}{1pt}
  \begin{tabular}{cccc}

    \includegraphics[width=0.25\textwidth]{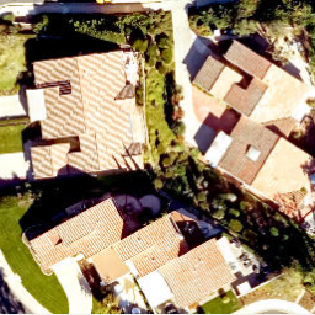}  & 
    \includegraphics[width=0.25\textwidth]{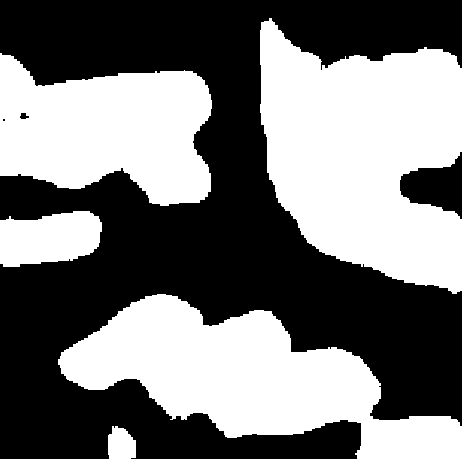} &
    \includegraphics[width=0.25\textwidth]{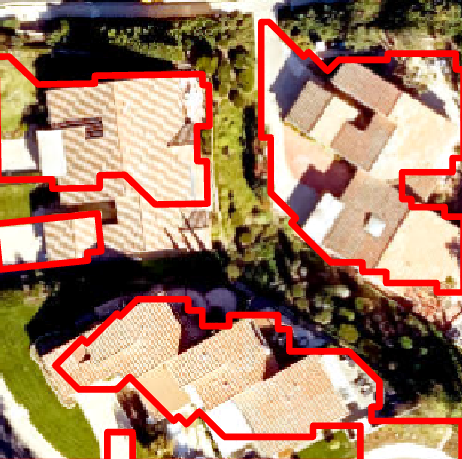} &
    \includegraphics[width=0.25\textwidth]{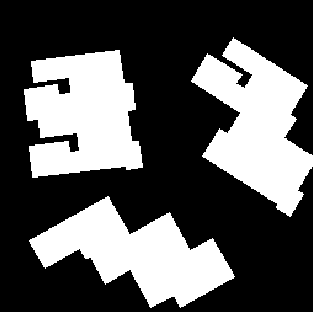} \\
  
    \includegraphics[width=0.25\textwidth]{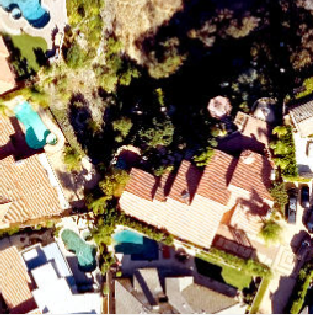}  & 
    \includegraphics[width=0.25\textwidth]{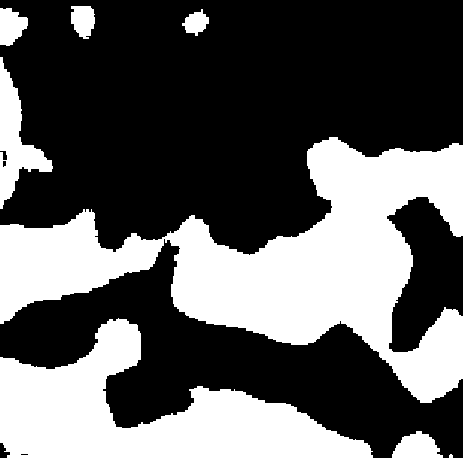} &
    \includegraphics[width=0.25\textwidth]{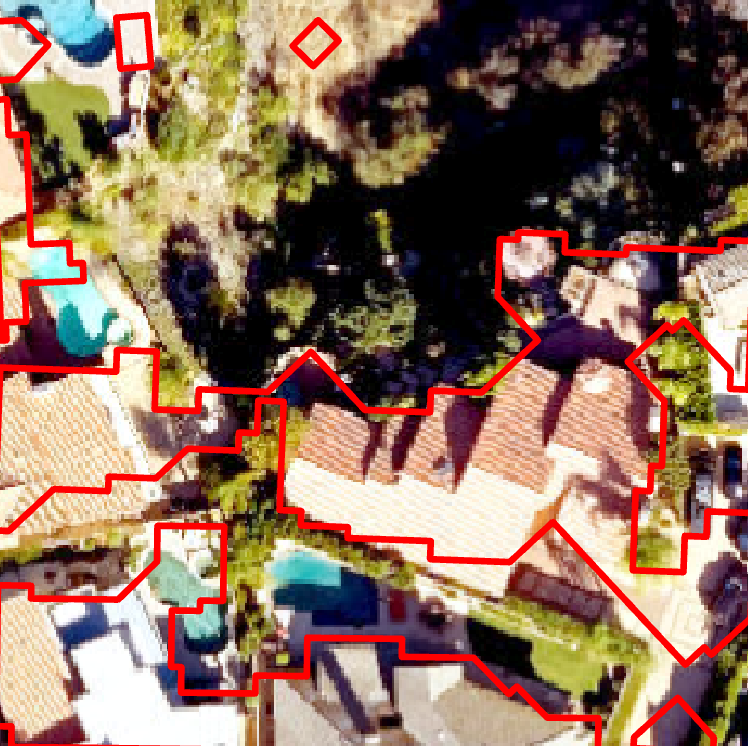} &
    \includegraphics[width=0.25\textwidth]{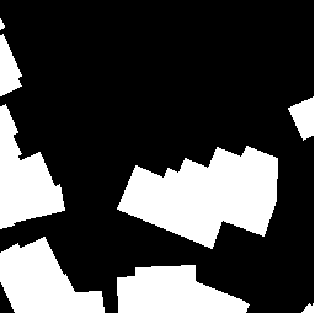} \\

    \includegraphics[width=0.25\textwidth]{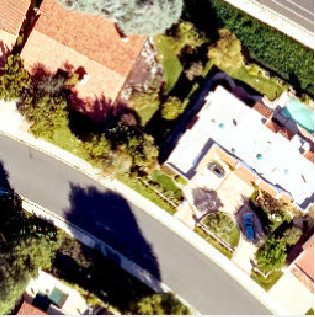}  & 
    \includegraphics[width=0.25\textwidth]{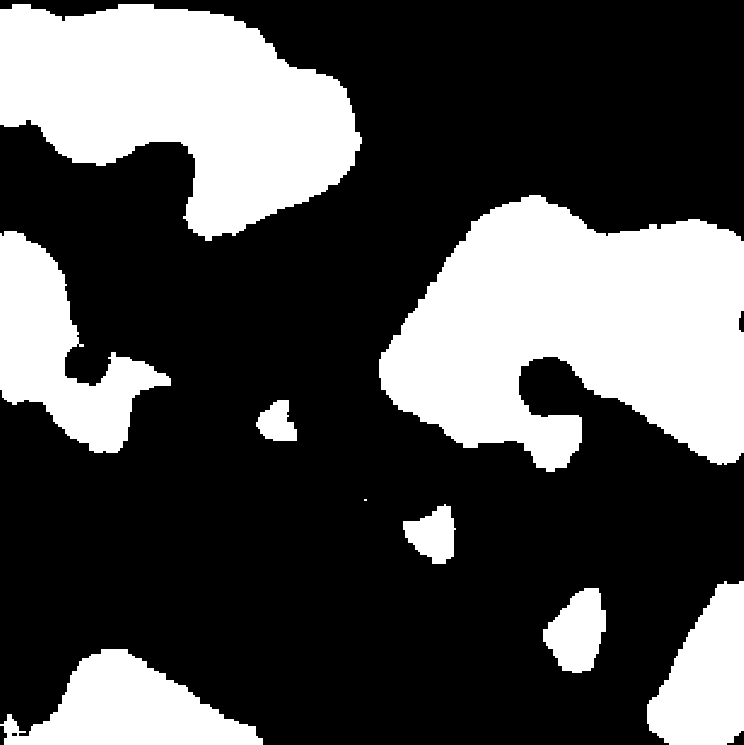} &
    \includegraphics[width=0.25\textwidth]{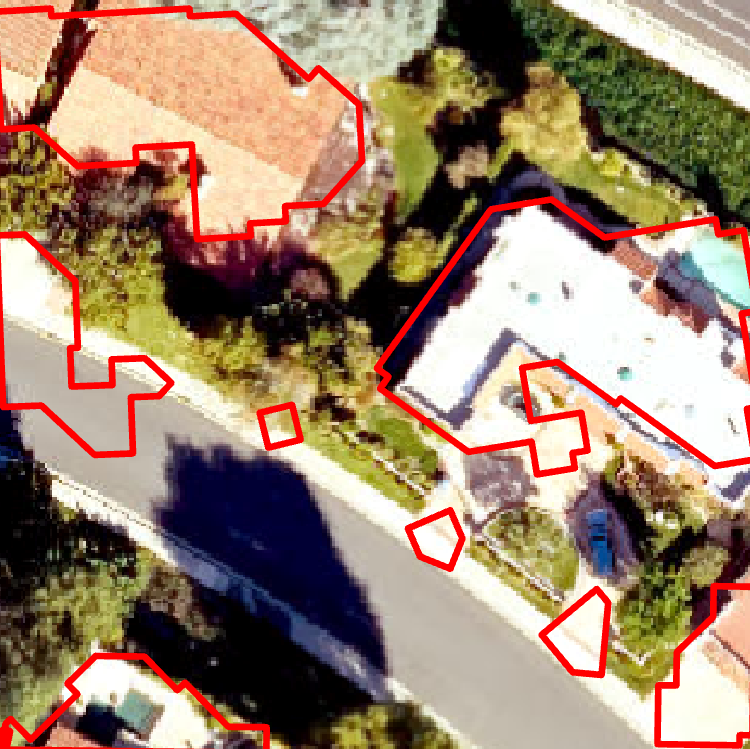} &
    \includegraphics[width=0.25\textwidth]{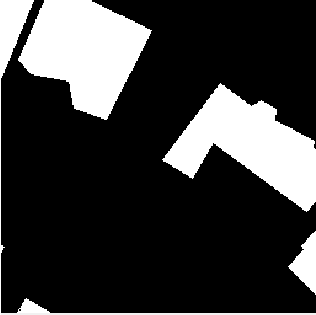} \\

    \includegraphics[width=0.25\textwidth]{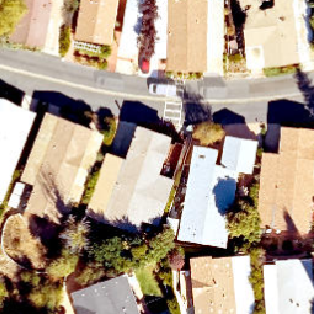}  & 
    \includegraphics[width=0.25\textwidth]{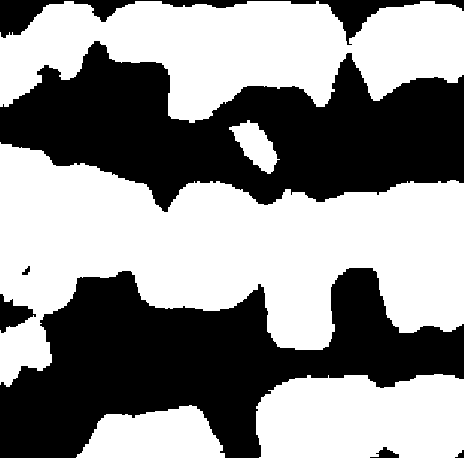} &
    \includegraphics[width=0.25\textwidth]{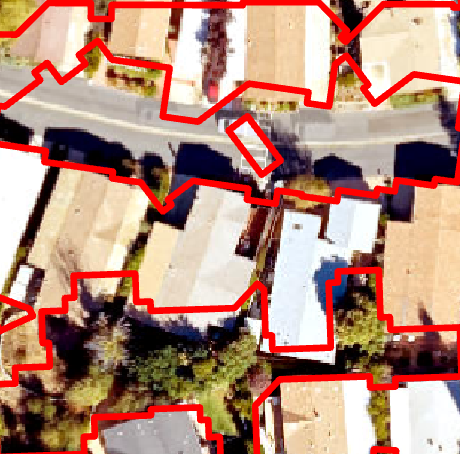} &
    \includegraphics[width=0.25\textwidth]{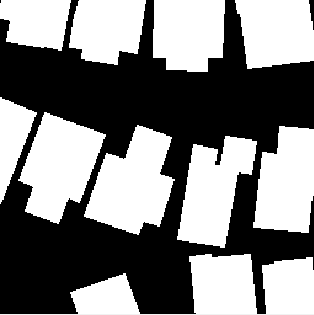} \\
    $I_R$ & $\hat{y}_{bin}$ & Regularized Footprint & $y$\\

  \end{tabular}

  \end{center}
  \caption[Output from the model $\hat{y}$ and the output after being post-processed 
  using the \texttt{Building Footprint Regularization} function.]{
    Binarized output from the model $\hat{y}_{bin}$ and the output after being post-processed 
    using the \texttt{Building Footprint Regularization} function. Please see 
    Section \ref{app:postprocessing} for more details.
  \label{fig:regularized_results}}
\end{figure}

\begin{figure}[!h]
  \begin{center}
  \setlength{\tabcolsep}{1pt}
  \begin{tabular}{cccccc}
    & $6000$ & $7000$ & $8000$ & $9000$ & $10000$ \\
   $\tilde{y}$ &
   \includegraphics[width=0.15\textwidth]{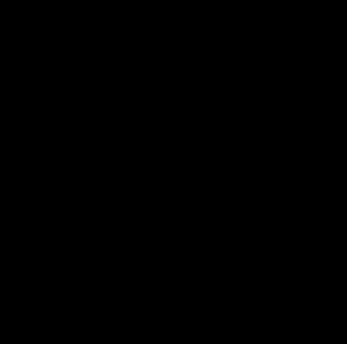} & 
   \includegraphics[width=0.15\textwidth]{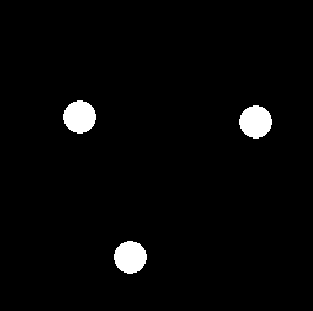} & 
   \includegraphics[width=0.15\textwidth]{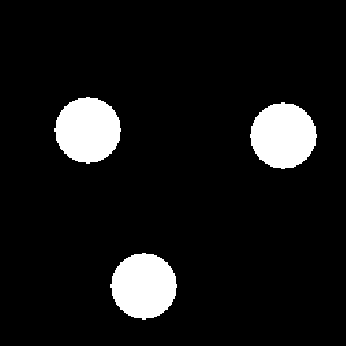} & 
   \includegraphics[width=0.15\textwidth]{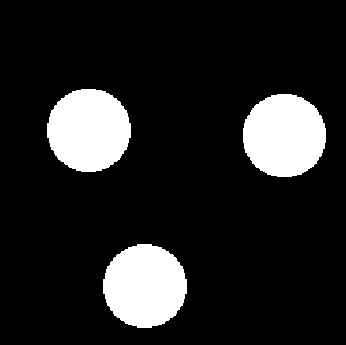} &
   \includegraphics[width=0.15\textwidth]{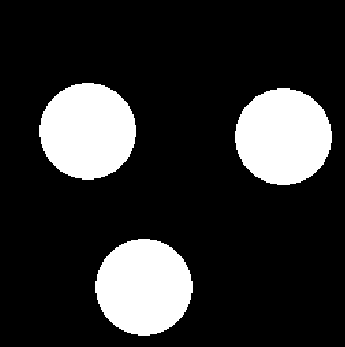} \\

   $\hat{y}$ &
   \includegraphics[width=0.15\textwidth]{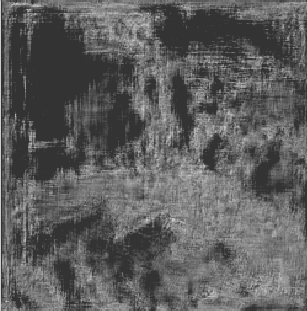} &
   \includegraphics[width=0.15\textwidth]{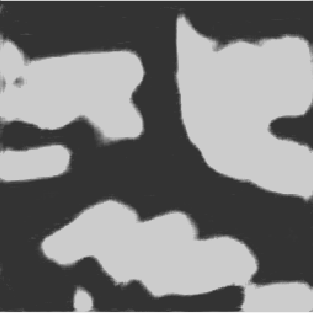} &
   \includegraphics[width=0.15\textwidth]{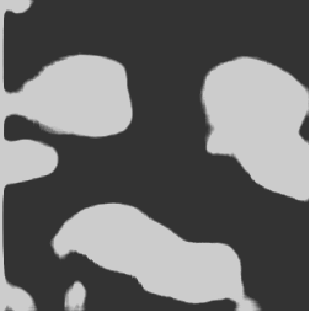} &
   \includegraphics[width=0.15\textwidth]{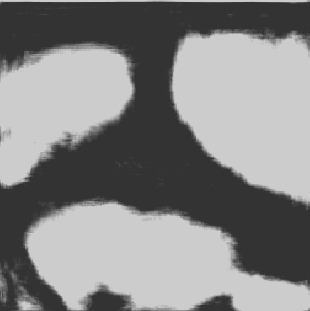} &
   \includegraphics[width=0.15\textwidth]{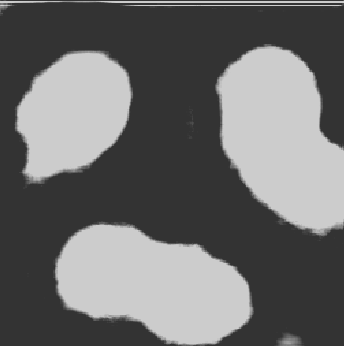} \\ 
  
   $I_{F_1}$ &
   \includegraphics[width=0.15\textwidth]{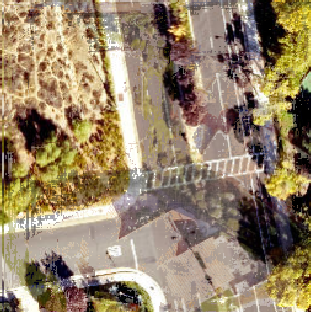} &
   \includegraphics[width=0.15\textwidth]{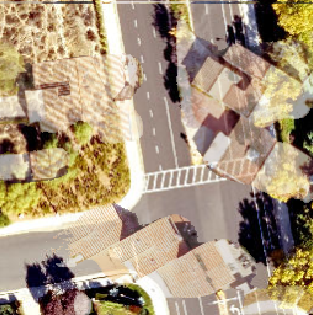} &
   \includegraphics[width=0.15\textwidth]{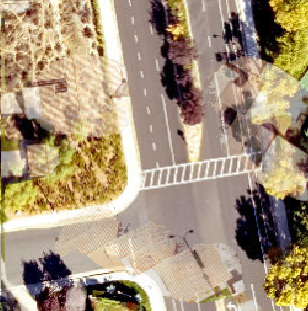} &
   \includegraphics[width=0.15\textwidth]{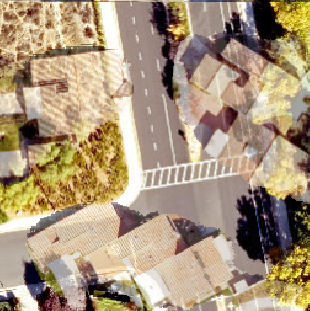} &
   \includegraphics[width=0.15\textwidth]{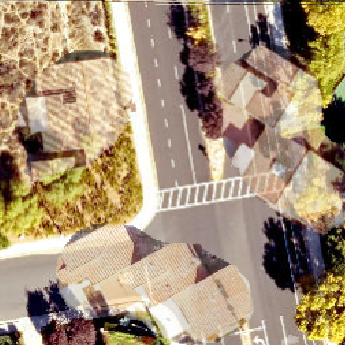} \\

   $I_{F_2}$ &
   \includegraphics[width=0.15\textwidth]{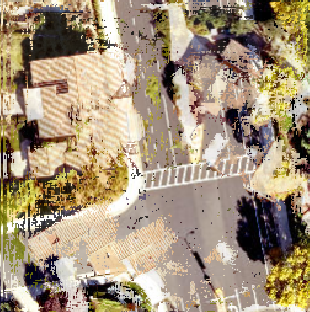} &
   \includegraphics[width=0.15\textwidth]{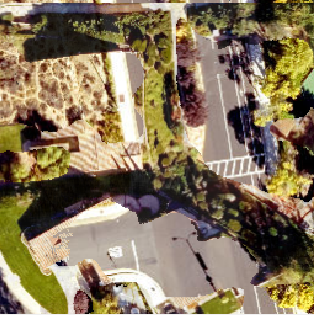} &
   \includegraphics[width=0.15\textwidth]{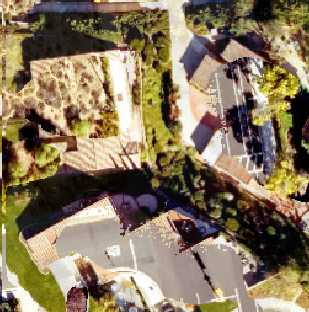} &
   \includegraphics[width=0.15\textwidth]{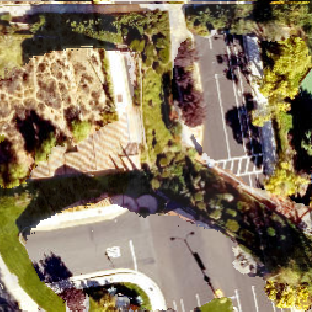} &
   \includegraphics[width=0.15\textwidth]{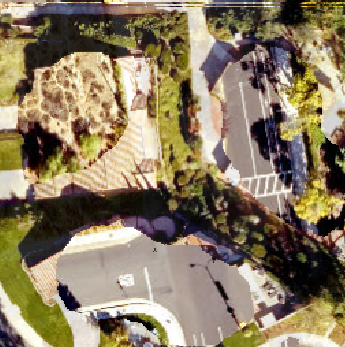} \\

  \end{tabular}\\

  \end{center}
  \caption[Visual results from the localization guidance ablative study for the Woolsey dataset.]{
    Visual results from the localization guidance ablative study for the Woolsey dataset, for different values 
    of csm.
  \label{fig:ablative_woolsey}}
  
\end{figure}

\begin{figure}[!h]
  \begin{center}
  \setlength{\tabcolsep}{1pt}
  \begin{tabular}{cccccc}
    & $6000$ & $7000$ & $8000$ & $9000$ & $10000$ \\
   $\tilde{y}$ &
   \includegraphics[width=0.15\textwidth]{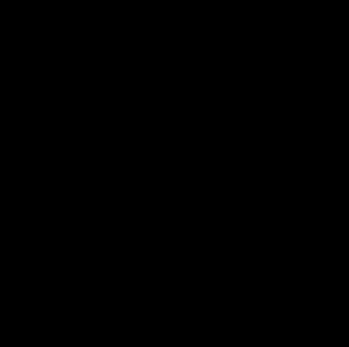}  & 
   \includegraphics[width=0.15\textwidth]{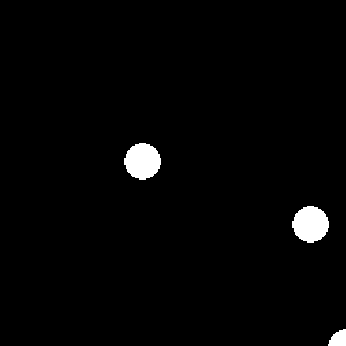} & 
   \includegraphics[width=0.15\textwidth]{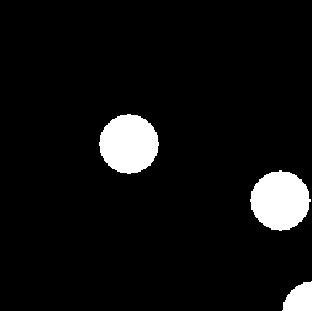}  & 
   \includegraphics[width=0.15\textwidth]{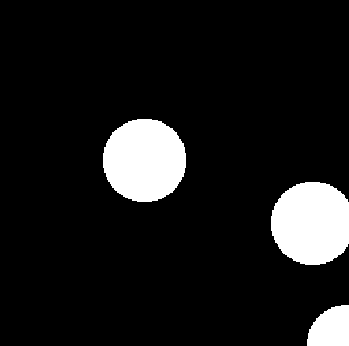}&
   \includegraphics[width=0.15\textwidth]{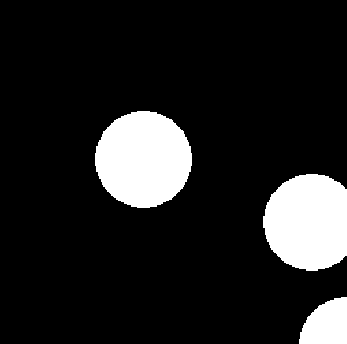} \\

   $\hat{y}$ &
   \includegraphics[width=0.15\textwidth]{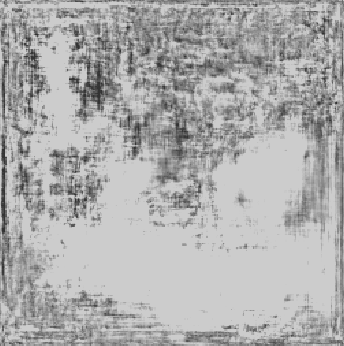} &
   \includegraphics[width=0.15\textwidth]{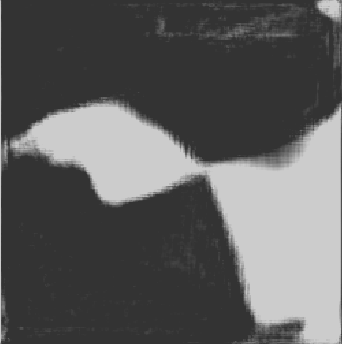} &
   \includegraphics[width=0.15\textwidth]{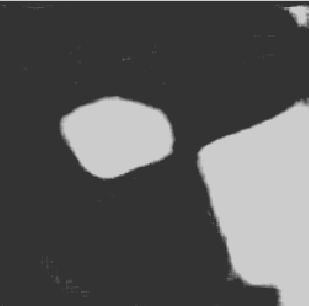} &
   \includegraphics[width=0.15\textwidth]{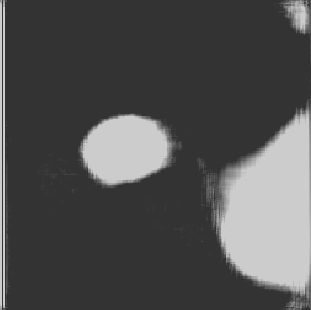} &
   \includegraphics[width=0.15\textwidth]{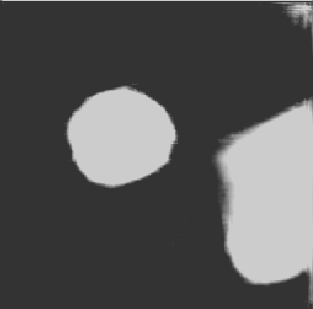} \\ 
  
   $I_{F_1}$ &
   \includegraphics[width=0.15\textwidth]{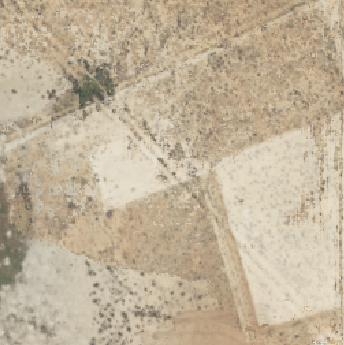} &
   \includegraphics[width=0.15\textwidth]{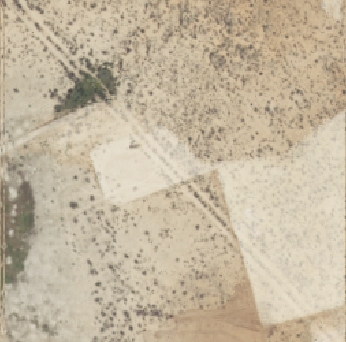} &
   \includegraphics[width=0.15\textwidth]{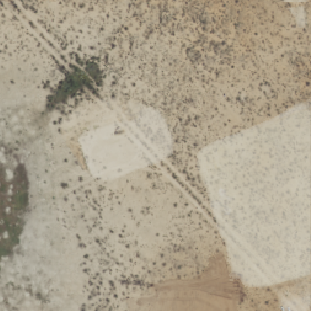} &
   \includegraphics[width=0.15\textwidth]{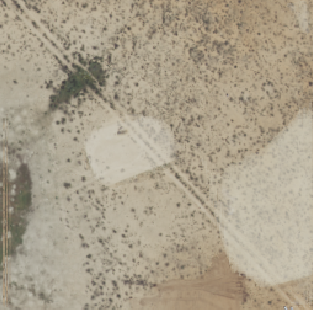} &
   \includegraphics[width=0.15\textwidth]{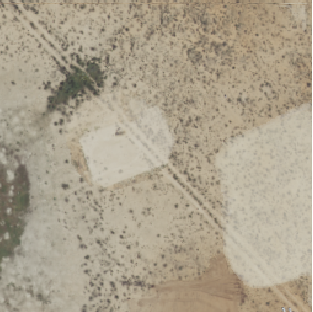} \\

   $I_{F_2}$ &
   \includegraphics[width=0.15\textwidth]{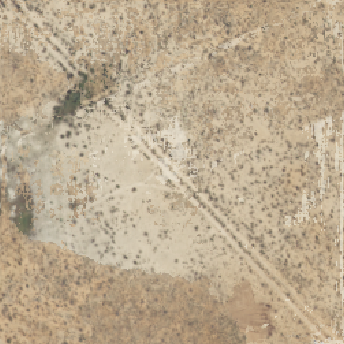} &
   \includegraphics[width=0.15\textwidth]{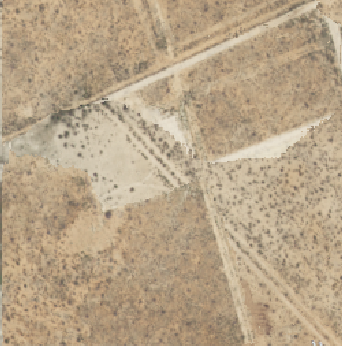} &
   \includegraphics[width=0.15\textwidth]{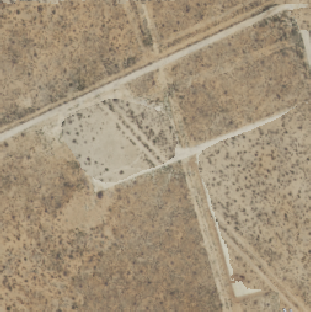} &
   \includegraphics[width=0.15\textwidth]{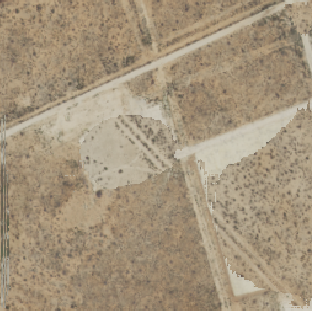} &
   \includegraphics[width=0.15\textwidth]{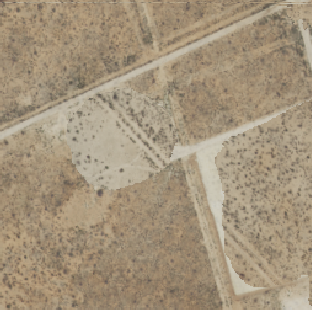} \\

  \end{tabular}\\

  \end{center}
  \caption[Visual results from the localization guidance ablative study for the Well Pads dataset.]{
    Visual results from the localization guidance ablative study for the Well Pads dataset, for different values 
    of csm. 
  \label{fig:ablative_wellpad}}
\end{figure}

\begin{figure}[!h]
  \begin{center}
  \setlength{\tabcolsep}{1pt}
  \begin{tabular}{cccccc}
    & $6000$ & $7000$ & $8000$ & $9000$ & $10000$ \\
   $\tilde{y}$ &
   \includegraphics[width=0.15\textwidth]{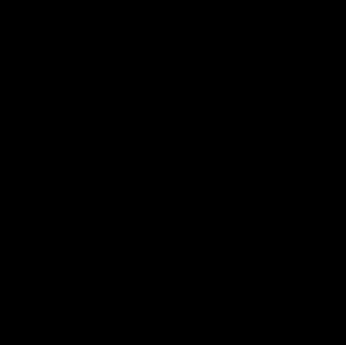}  & 
   \includegraphics[width=0.15\textwidth]{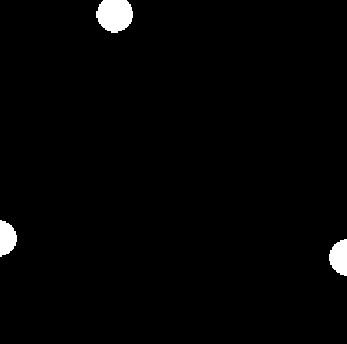} & 
   \includegraphics[width=0.15\textwidth]{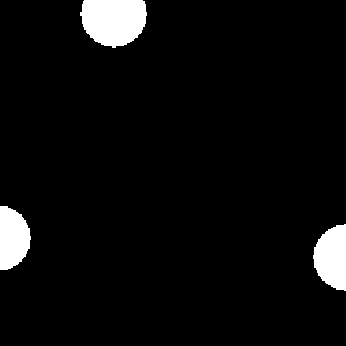} & 
   \includegraphics[width=0.15\textwidth]{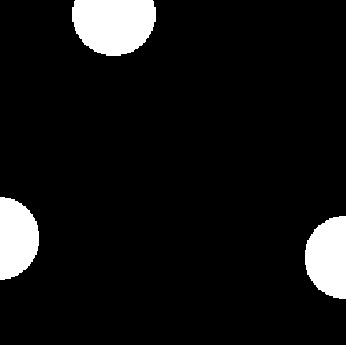} &
   \includegraphics[width=0.15\textwidth]{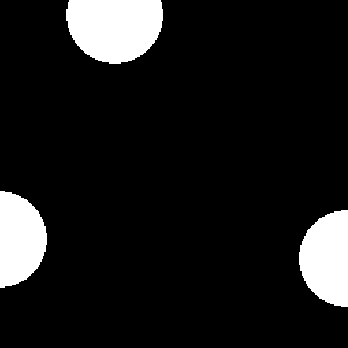} \\

   $\hat{y}$ &
   \includegraphics[width=0.15\textwidth]{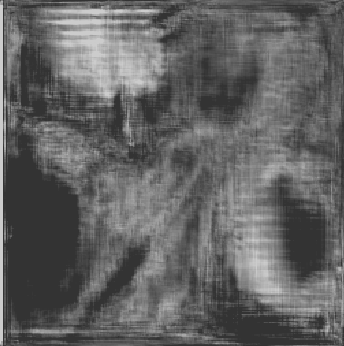} &
   \includegraphics[width=0.15\textwidth]{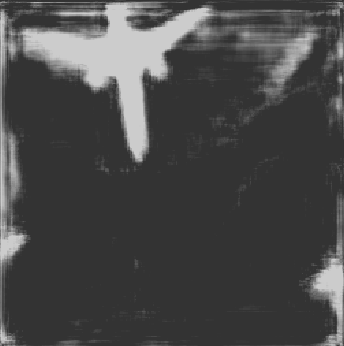} &
   \includegraphics[width=0.15\textwidth]{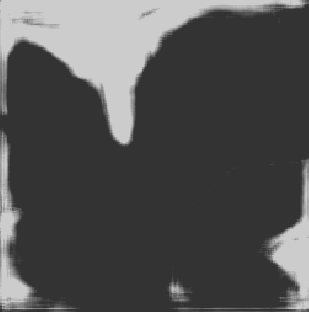} &
   \includegraphics[width=0.15\textwidth]{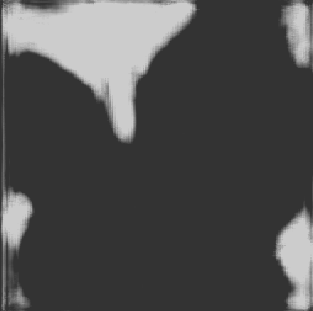} &
   \includegraphics[width=0.15\textwidth]{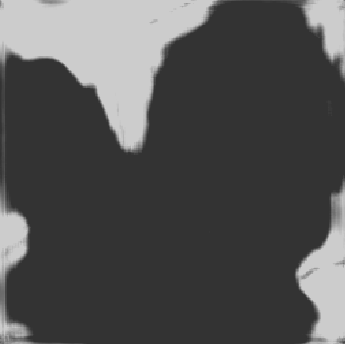} \\ 
  
   $I_{F_1}$ &
   \includegraphics[width=0.15\textwidth]{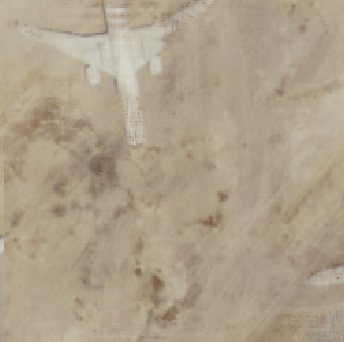} &
   \includegraphics[width=0.15\textwidth]{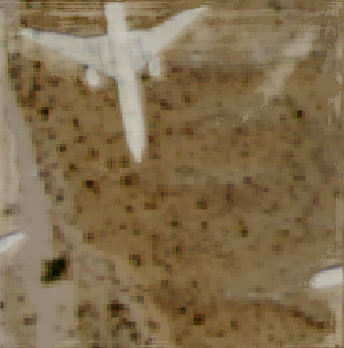} &
   \includegraphics[width=0.15\textwidth]{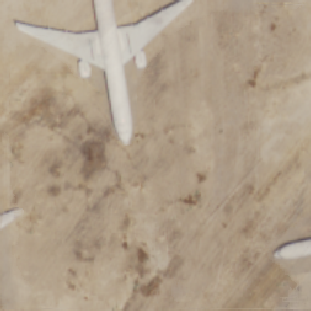} &
   \includegraphics[width=0.15\textwidth]{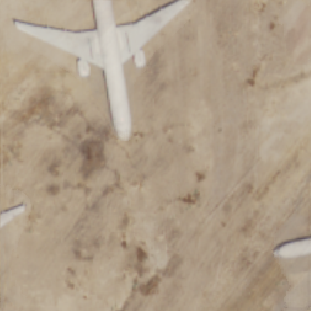} &
   \includegraphics[width=0.15\textwidth]{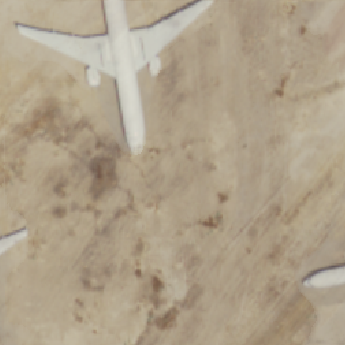} \\

   $I_{F_2}$ &
   \includegraphics[width=0.15\textwidth]{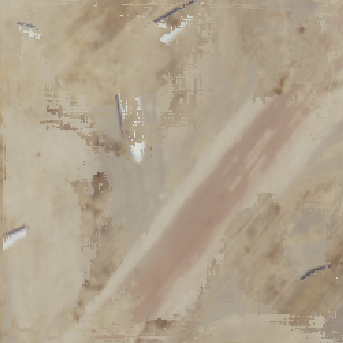} &
   \includegraphics[width=0.15\textwidth]{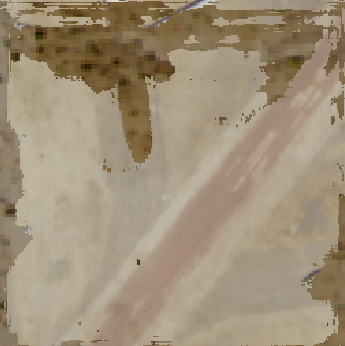} &
   \includegraphics[width=0.15\textwidth]{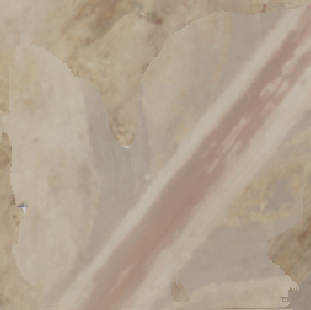} &
   \includegraphics[width=0.15\textwidth]{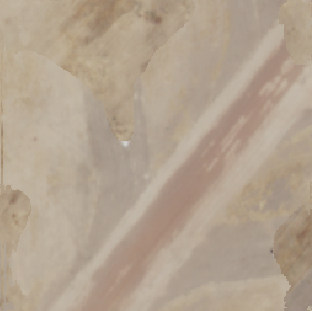} &
   \includegraphics[width=0.15\textwidth]{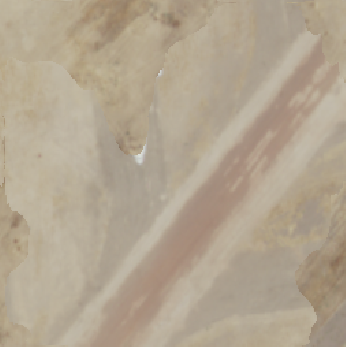} \\

  \end{tabular}\\

  \end{center}
  \caption[Visual results from the localization guidance ablative study for the Aircraft dataset.]{
    Visual results from the localization guidance ablative study for the Aircraft dataset, for different values 
    of csm.
  \label{fig:ablative_aircraft}}
\end{figure}

%
\begin{figure}[!h]
  \begin{center}
  \setlength{\tabcolsep}{1pt}
  \begin{tabular}{ccccccc}
    & Input $I_R$ & Pseudo label $\tilde{y}$ & Output $\hat{y}$ & Pos. Fake $I_{F_1}$ & Neg. Fake $I_{F_2}$ & Ground Truth $y$ \\
    \multirow{2}{*}{\rotatebox[origin=c]{90}{Airplanes}} & 
    
    \includegraphics[width=0.15\textwidth]{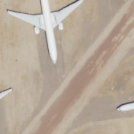} & 
   \includegraphics[width=0.15\textwidth]{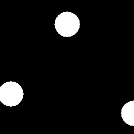} & 
   \includegraphics[width=0.15\textwidth]{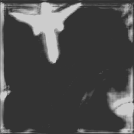} &
   \includegraphics[width=0.15\textwidth]{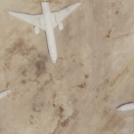} &
   \includegraphics[width=0.15\textwidth]{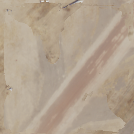} &
   \includegraphics[width=0.15\textwidth]{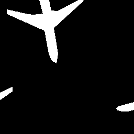} \\ 
   &
   \includegraphics[width=0.15\textwidth]{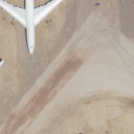} &
   \includegraphics[width=0.15\textwidth]{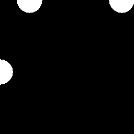} & 
   \includegraphics[width=0.15\textwidth]{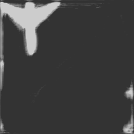} &
   \includegraphics[width=0.15\textwidth]{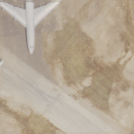} &
   \includegraphics[width=0.15\textwidth]{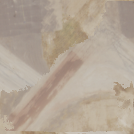} &
   \includegraphics[width=0.15\textwidth]{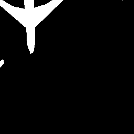} \\
    
   \multirow{2}{*}{\rotatebox[origin=c]{90}{Well Pad}} &
   \includegraphics[width=0.15\textwidth]{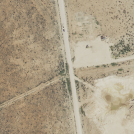} &
   \includegraphics[width=0.15\textwidth]{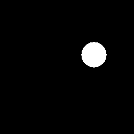} & 
   \includegraphics[width=0.15\textwidth]{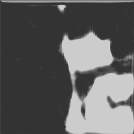} &
   \includegraphics[width=0.15\textwidth]{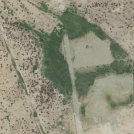} &
   \includegraphics[width=0.15\textwidth]{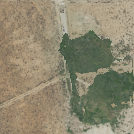} &
   \includegraphics[width=0.15\textwidth]{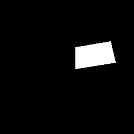} \\
   & \includegraphics[width=0.15\textwidth]{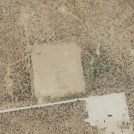} &
   \includegraphics[width=0.15\textwidth]{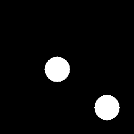} & 
   \includegraphics[width=0.15\textwidth]{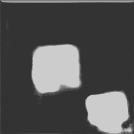} &
   \includegraphics[width=0.15\textwidth]{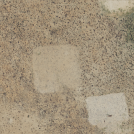} &
   \includegraphics[width=0.15\textwidth]{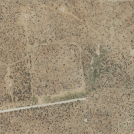} &
   \includegraphics[width=0.15\textwidth]{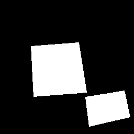} \\

   \multirow{2}{*}{\rotatebox[origin=c]{90}{Woolsey}} &
   \includegraphics[width=0.15\textwidth]{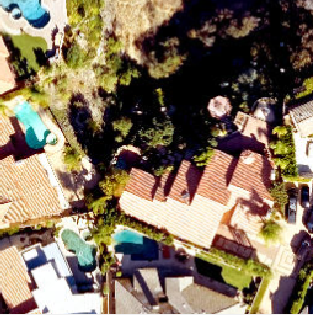} &
   \includegraphics[width=0.15\textwidth]{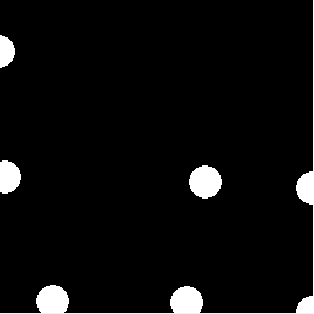} & 
   \includegraphics[width=0.15\textwidth]{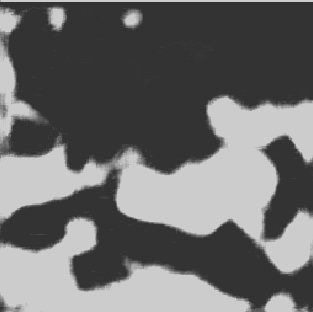} &
   \includegraphics[width=0.15\textwidth]{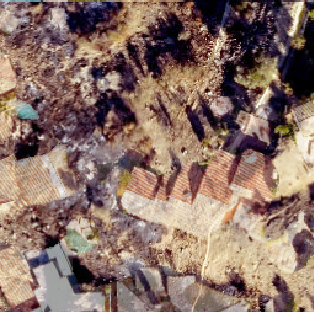} &
   \includegraphics[width=0.15\textwidth]{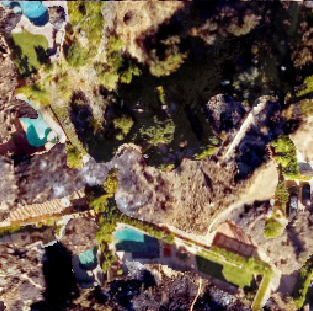} &
   \includegraphics[width=0.15\textwidth]{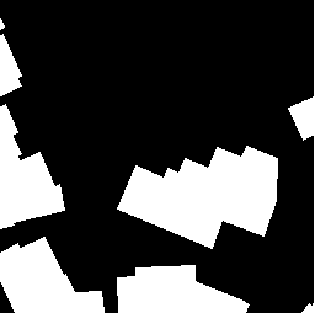} \\
   & 
   \includegraphics[width=0.15\textwidth]{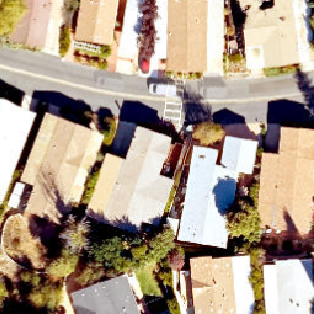} &
   \includegraphics[width=0.15\textwidth]{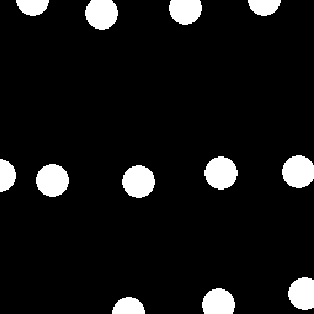} & 
   \includegraphics[width=0.15\textwidth]{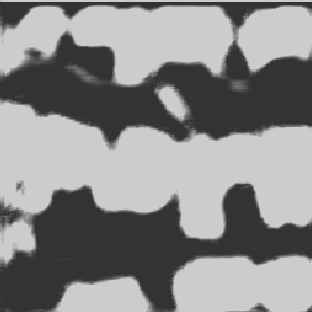} &
   \includegraphics[width=0.15\textwidth]{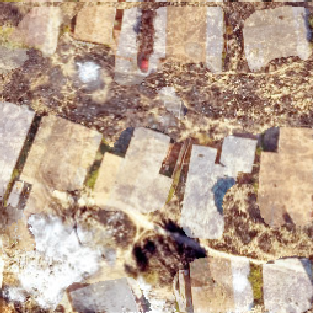} &
   \includegraphics[width=0.15\textwidth]{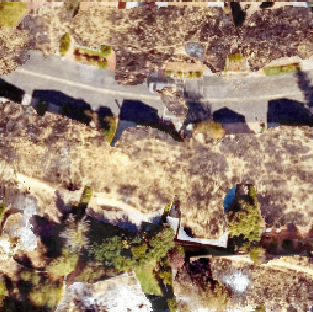} &
   \includegraphics[width=0.15\textwidth]{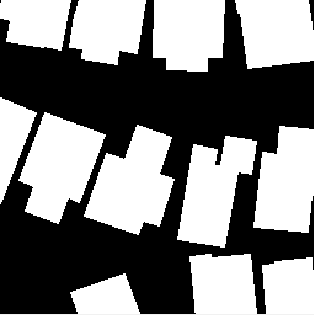} \\
   \multirow{2}{*}{\rotatebox[origin=c]{90}{SpaceNet}} &
   \includegraphics[width=0.15\textwidth]{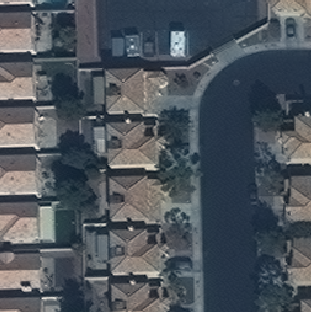} &
   \includegraphics[width=0.15\textwidth]{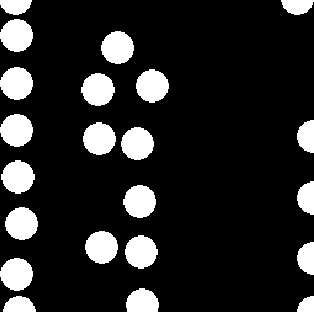} & 
   \includegraphics[width=0.15\textwidth]{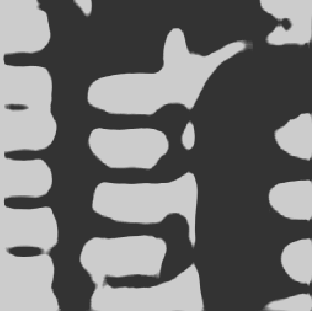} &
   \includegraphics[width=0.15\textwidth]{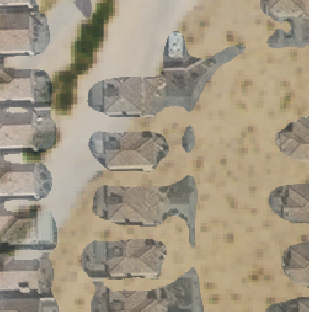} &
   \includegraphics[width=0.15\textwidth]{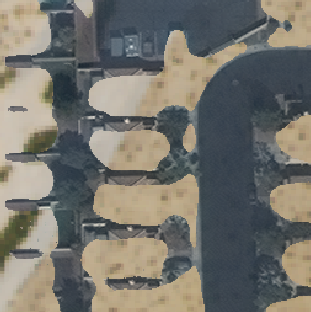} &
   \includegraphics[width=0.15\textwidth]{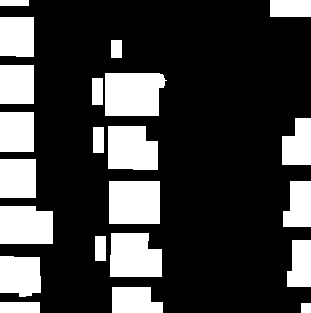} \\
   & 
   \includegraphics[width=0.15\textwidth]{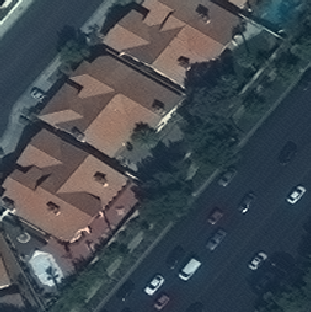} &
   \includegraphics[width=0.15\textwidth]{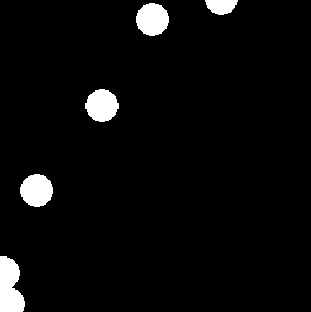} & 
   \includegraphics[width=0.15\textwidth]{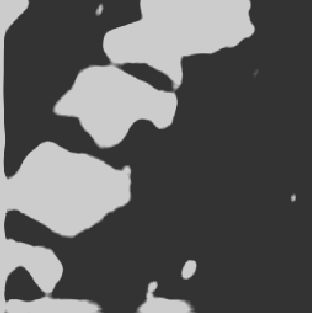} &
   \includegraphics[width=0.15\textwidth]{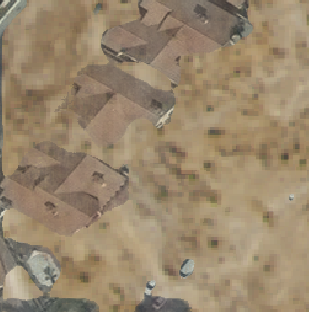} &
   \includegraphics[width=0.15\textwidth]{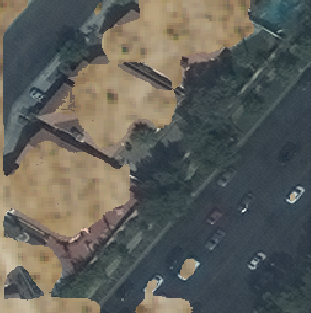} &
   \includegraphics[width=0.15\textwidth]{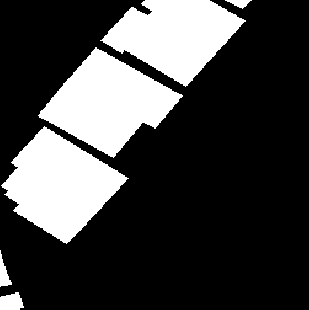} \\
   
  \end{tabular}
  \end{center}
  \caption[Example images and segmentation masks for two samples in each dataset.]{
    Example segmentation masks and training data for two samples in each dataset. 
  \label{fig:airplane_results}}
  
\end{figure}

\end{document}